\documentclass[10pt,twocolumn,letterpaper]{article}

\usepackage{wacv}
\usepackage{times}
\usepackage{epsfig}
\usepackage{graphicx}
\usepackage{amsmath}
\usepackage{amssymb}
\usepackage{multirow,tabularx}
\usepackage{xcolor}
\usepackage{soul}
\usepackage{booktabs}
\usepackage{pifont}

\usepackage[font=small]{caption}

\usepackage{subcaption}

\newcommand{\hlc}[2][yellow]{{%
    \colorlet{foo}{#1}%
    \sethlcolor{foo}\hl{#2}}%
}

\newcommand{\cmark}{\ding{51}}%
\newcommand{\xmark}{{\color{black} \ding{55}}}%

\newcommand{\datasetname}{InfographicVQA\xspace}
\newcommand{\docvqa}{DocVQA\xspace}

\newcommand{\textvqa}{TextVQA\xspace}
\newcommand{\stvqa}{ST-VQA\xspace}
\newcommand{\ocrvqa}{OCR-VQA\xspace}

\newcommand{\visualmrc}{VisualMRC\xspace}
\newcommand{\drop}{DROP\xspace}
\newcommand{\dvqa}{DVQA\xspace}
\newcommand{\leafqa}{LEAF-QA\xspace}
\newcommand{\figureqa}{FigureQA\xspace}
\newcommand{\textbookqa}{TQA\xspace}
\newcommand{\recipeqa}{RecipeQA\xspace}

\newcommand{\visuallydataset}{Visually29K\xspace}

\newcommand{\uniter}{UNITER\xspace}
\newcommand{\visualbert}{VisualBERT\xspace}

\newcommand{\vlbert}{VL-BERT\xspace}

\newcommand{\layoutlm}{LayoutLM\xspace}
\newcommand{\lambert}{LAMBERT\xspace}

\newcommand{\correct}[1]{\textcolor{green}{#1}}
\newcommand{\halfcorrect}[1]{\textcolor{orange}{#1}}
\newcommand{\wrong}[1]{\textcolor{red}{#1}}
\newcommand{\groundtruth}[1]{\textcolor{blue}{\textbf{#1}}}

\usepackage{hyperref}
\hypersetup{breaklinks=true,colorlinks}

\wacvfinalcopy %

\begin{document}

\title{InfographicVQA}

\author
{Minesh Mathew$^{1}$ $\qquad$ Viraj Bagal$^{2}$\thanks{Work done during an internship at IIIT Hyderabad.} $\qquad$ Rub{\`e}n Tito$^{3}$ \\
Dimosthenis Karatzas$^{3}$  $\qquad$ Ernest Valveny$^{3}$ $\qquad$   C.V. Jawahar$^{1}$ \\
$^1$CVIT, IIIT Hyderabad, India $\qquad$ 
$^1$IISER Pune, India $\qquad$ 
$^3$CVC, Universitat Autònoma de Barcelona, Spain$\qquad$
\\
{\tt\small minesh.mathew@research.iiit.ac.in, viraj.bagal@students.iiserpune.ac.in, rperez@cvc.uab.cat}
}

\maketitle

\begin{abstract}
Infographics communicate information using a combination of textual, graphical and visual elements.
In this work, we explore automatic understanding of infographic images by using a Visual Question Answering technique.
To this end, we present \datasetname, a new dataset that comprises a diverse collection of infographics along with  question-answer annotations. The questions require methods to jointly reason over the document layout, textual content, graphical elements, and data visualizations. We curate the dataset with emphasis on questions that require elementary reasoning and basic arithmetic skills. For VQA on the dataset, we evaluate two strong baselines based on state-of-the-art Transformer-based, scene text VQA and document understanding models. 
Poor performance of  both the  approaches compared to near perfect human performance suggests that VQA on infographics that are designed to communicate information quickly and clearly to human brain, is ideal for benchmarking machine understanding of complex document images. The dataset, code and leaderboard will be made available at \href{http://docvqa.org}{docvqa.org}

\end{abstract}

\vspace{-3mm}
\section{Introduction}
\label{sec:intro}
Infographics are documents created to convey information in a compact manner using a combination of textual and visual cues.
The presence of the text, numbers and symbols, along with the semantics that arise from their relative placements, make infographics understanding a challenging problem.
True document image understanding in this domain requires methods to jointly reason over the document layout, textual content, graphical elements, data visualisations, color schemes and visual art among others.
Motivated by the multimodal nature of infographics, and the human centred design, we propose a Visual Question Answering (VQA) approach to infographics understanding.  

VQA received significant attention over the past few years~\cite{vqa2,anderson2018bottom,gurari2018vizwiz,gqa,saaa,san}. 
Several new VQA branches focus on images with text, such as answering questions by looking at text books~\cite{textbookqa}, business documents~\cite{docvqa}, charts~\cite{dvqa,figureqa,leafqa} and screenshots of web pages~\cite{visualmrc}.
Still, infographics are unique in their combined use, and purposeful arrangement of visual and textual elements.

\begin{figure}[t]
\small

\begin{center}
\begin{tabular}{p{0.95\linewidth}}
\includegraphics[width=\linewidth,height=0.65\linewidth]{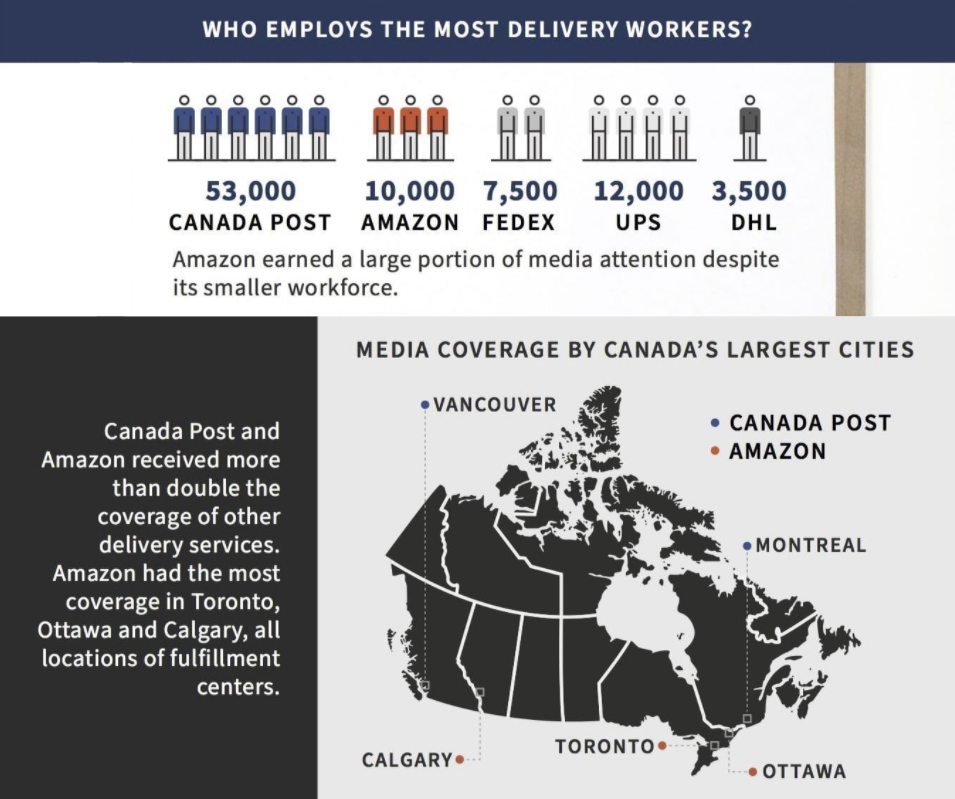}   \\
How many companies have more than 10K delivery workers?
Answer: {2} \hspace{2.73cm}
Evidence:  \hlc[cyan!50]{Figure}  \par
Answer-source:  \hlc[blue!30]{Non-extractive} \hspace{0.1cm}
Operation:  \hlc[yellow!50]{Counting} \hspace{0.05cm} \hlc[yellow!50]{Sorting} \\

\midrule

Who has better coverage in Toronto - Canada post or Amazon? \par
Answer: {canada post} \hspace{2.95cm}
Evidence: \hlc[cyan!50]{Text} \par
Answer-source: \hlc[blue!30]{Question-span}  
\hspace{0.05cm} \hlc[blue!30]{Image-span}
\hspace{0.05cm}
Operation: {\textit{none}}  \\

\midrule

In which cities did Canada Post get maximum media coverage? \par
Answer: {vancouver, montreal} \hspace{0.2cm} \hspace{0.75cm}
Evidence:  \hlc[cyan!50]{Text} \hspace{0.05cm}  \hlc[cyan!50]{Map} \par 
Answer-source:  \hlc[blue!30]{Multi-span} \hspace{1.3cm}
 Operation: {\textit{none}}  \\

\end{tabular}
\vspace{-3mm}
\caption{Example image from \datasetname along with   questions and answers. For each question,  source of the answer, type of evidence the answer is grounded on, and the discrete operation required to find the answer are shown. }
\label{fig:intro_fig}
\end{center}
\end{figure}

\newcommand{\machinereadable}{MR}
\newcommand{\scenetext}{ST}
\newcommand{\borndigital}{BD}
\newcommand{\printed}{Pr}
\newcommand{\typewritten}{Tw}
\newcommand{\handwritten}{Hw}

\newcommand{\multchoicequest}{MCQ}
\newcommand{\extractive}{Ex}
\newcommand{\shortabstractive}{SAb}
\newcommand{\abstractive}{Ab}
\newcommand{\yesno}{Y/N}
\newcommand{\numerical}{Nm}

\begin{table*}[ht]
\footnotesize
\begin{center}
\begin{tabular}{llcclrrl}
\toprule
\multirow{2}{*}{Dataset} & \multirow{2}{*}{Images} & Synthetic & Template  & \multirow{2}{*}{Text type} & \multirow{2}{*}{\# Images} & \multirow{2}{*}{\# Questions} & \multirow{2}{*}{Answer type} \\
 & & Images & questions & & & & \\
\midrule
\textbookqa~\cite{textbookqa} & Science diagrams    & \xmark & \xmark & \machinereadable & 1K     & 26K   & \multchoicequest                \\
\recipeqa~\cite{recipeqa}     & Culinary pictures     & \xmark & \cmark & \machinereadable & 251K   & 37K   & \multchoicequest                \\
\stvqa~\cite{stvqa_iccv}      & Natural images     & \xmark & \xmark & \scenetext       & 23K    & 31K   & \extractive                     \\
\textvqa~\cite{textvqa}       & Natural images      & \xmark & \xmark & \scenetext       & 28K    & 45K   & \extractive, \shortabstractive  \\
\ocrvqa~\cite{ocr-vqa}        & Book covers        & \xmark & \cmark & \borndigital     & 207K   & 1M    & \extractive, \yesno             \\
\dvqa~\cite{dvqa}             & Bar charts         & \cmark & \cmark & \borndigital     & 300K   & 3.4M & \extractive, \numerical, \yesno  \\
\figureqa~\cite{figureqa}     & Charts - 5 types   & \cmark & \cmark & \borndigital     & 120K   & 1.5M  & \yesno                          \\
\leafqa~\cite{leafqa}         & Charts - 4 types   & \cmark & \cmark & \borndigital     & 250K   & 2M    & \extractive, \numerical, \yesno \\
\visualmrc~\cite{visualmrc}   & Webpage screenshots     & \xmark & \xmark & \borndigital     & 10K    & 30K   & \abstractive                    \\
\docvqa~\cite{docvqa}         & Industry documents      & \xmark & \xmark & \printed, \typewritten, \handwritten, \borndigital & 12K & 50K & \extractive \\
\datasetname                  & Infographics       & \xmark & \xmark & \borndigital     & 5.4K   & 30K   & \extractive, \numerical         \\         
\bottomrule
\end{tabular}
\vspace{-3mm}
\caption{\textbf{Summary of VQA and Multimodal QA datasets where text on the images need to be read to answer questions.} Text type abbreviations are: Machine Readable: MR, Scene Text: ST,Born Digital: BD, Printed: Pr, Handwritten: Hw, and  Typewritten: Tw. Answer type abbreviations are: Multiple Choice Question: MCQ, Extractive: Ex, Short abstractive: SAb, Abstractive: Ab, Yes/No: Y/N, and  Numerical (answer is numerical and not extracted from image or question; but derived): Nm.}
\label{tab:dataset_statistics}
\end{center}
\end{table*}

In this work, we introduce a new dataset for VQA on infographics, \datasetname,  comprising $30,035$ questions over $5,485$ images. An example from our dataset is shown in~\autoref{fig:intro_fig}. 
Questions in the dataset include questions grounded on tables, figures and visualizations as well as questions that require combining multiple cues. Since most infographics contain numerical data, we  collect questions that require elementary reasoning skills such as counting, sorting and arithmetic operations.
We believe our dataset is ideal for benchmarking progress of algorithms at the meeting point of vision, language and document understanding.

We adapt a multimodal Transformer~\cite{attention_is_all_you_need}-based VQA model called M4C~\cite{m4c} and a layout-aware, BERT~\cite{bert}-style extractive QA model called LayoutLM~\cite{layoutlm1} for VQA on \datasetname.
Results using these two strong baselines show that current state-of-the-art (SoTA) models for similar tasks perform poorly on the new dataset. The results also highlight the need to devise better feature extractors for infographics, different from bottom-up features~\cite{topdown_bottomup} that are typically used for VQA on natural scene images.

\vspace{-2mm}
\section{Related works}
\label{sec:related_works}

\noindent\textbf{Question answering in a multimodal context.}
Textbook Question Answering (\textbookqa)~\cite{textbookqa} and \recipeqa~\cite{recipeqa} are two works addressing Question Answering (QA) in a multimodal context.
For \textbookqa, contexts are textbook lessons and for \recipeqa, contexts are recipes containing text and images.
Contrary to \datasetname and other datasets mentioned below, text
in these two datasets are not embedded on the images, but provided in machine readable form, as a separate input.

\stvqa~\cite{stvqa_iccv} and \textvqa~\cite{textvqa} datasets extend VQA over natural images to a new direction where understanding scene text on the images is necessary to answer the questions.
While these datasets comprise images captured in the wild with sparse text content, \datasetname has born-digital images with an order of magnitude more text tokens per image, richer in layout and in the interplay between textual and visual elements.
\ocrvqa~\cite{ocr-vqa} introduces a task similar to \stvqa and \textvqa, but solely on images of book covers. %
Template questions are generated from book metadata  %
such as author name, title and other information. Consequently, questions in this dataset rely less %
on visual information present in the images.
\dvqa~\cite{dvqa}, \figureqa~\cite{figureqa}, and \leafqa~\cite{leafqa} datasets deal with VQA on charts. 
All the three datasets have chart images rendered using chart plotting libraries and template questions.  

\docvqa~\cite{docvqa} is  a VQA dataset that comprises  document images of industry/business documents, and questions that require understanding document elements such as text passages, forms, and tables.
Similar to \stvqa, \docvqa is an extractive VQA task where answers can always be extracted verbatim from the text on the images.
\visualmrc~\cite{visualmrc} on the other hand   is an abstractive VQA (answers cannot  be directly extracted from text in the images or questions) benchmark where images are screenshots of web pages. 
Compared to \visualmrc, \datasetname is an extractive VQA task (answers are extracted as `span'(s) of the question or text present in the given image), except for questions that require certain discrete operations resulting in numerical non-extractive answers. (see~\autoref{subsec:types_evidence_operation}).
In~\autoref{tab:dataset_statistics} we present a high level summary of the QA/VQA datasets related to ours.

\textbf{Multimodal transformer for Vision-Language tasks.} Following the success of BERT~\cite{bert}-like models for Natural Language Processing (NLP) tasks, there have been multiple works extending it to the Vision-Language space. Models like \vlbert~\cite{vlbert}, \visualbert~\cite{visualbert}, and \uniter~\cite{uniter} show that combined pretraining of BERT-like architectures on vision and language inputs achieve SoTA performances on various downstream tasks including VQA on natural images.
For VQA on images with scene text,  M4C and TAP~\cite{tap} use a multimodal transformer block to fuse embeddings of question, scene text tokens, and objects detected from an image.
\begin{figure*}[t]
    \small
    \centering
    \begin{subfigure}[t]{0.32\textwidth}
        \includegraphics[width=1\linewidth]{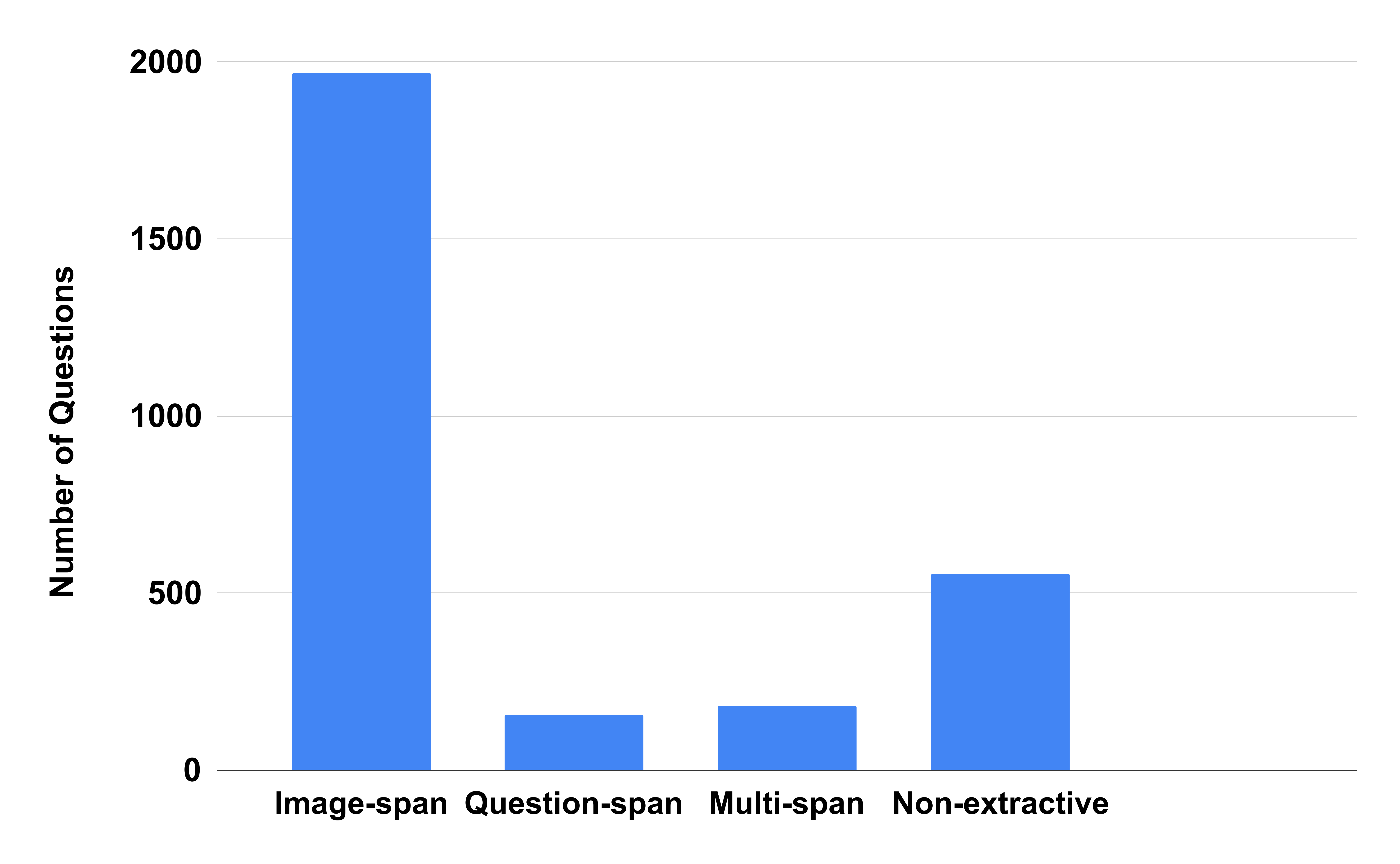}
        \vspace{-5mm}
        \caption{Answer-sources and their counts}
        \label{fig:answer_types}
    \end{subfigure}
    \hspace{1mm}
    \begin{subfigure}[t]{0.32\textwidth}
        \includegraphics[width=1\linewidth]{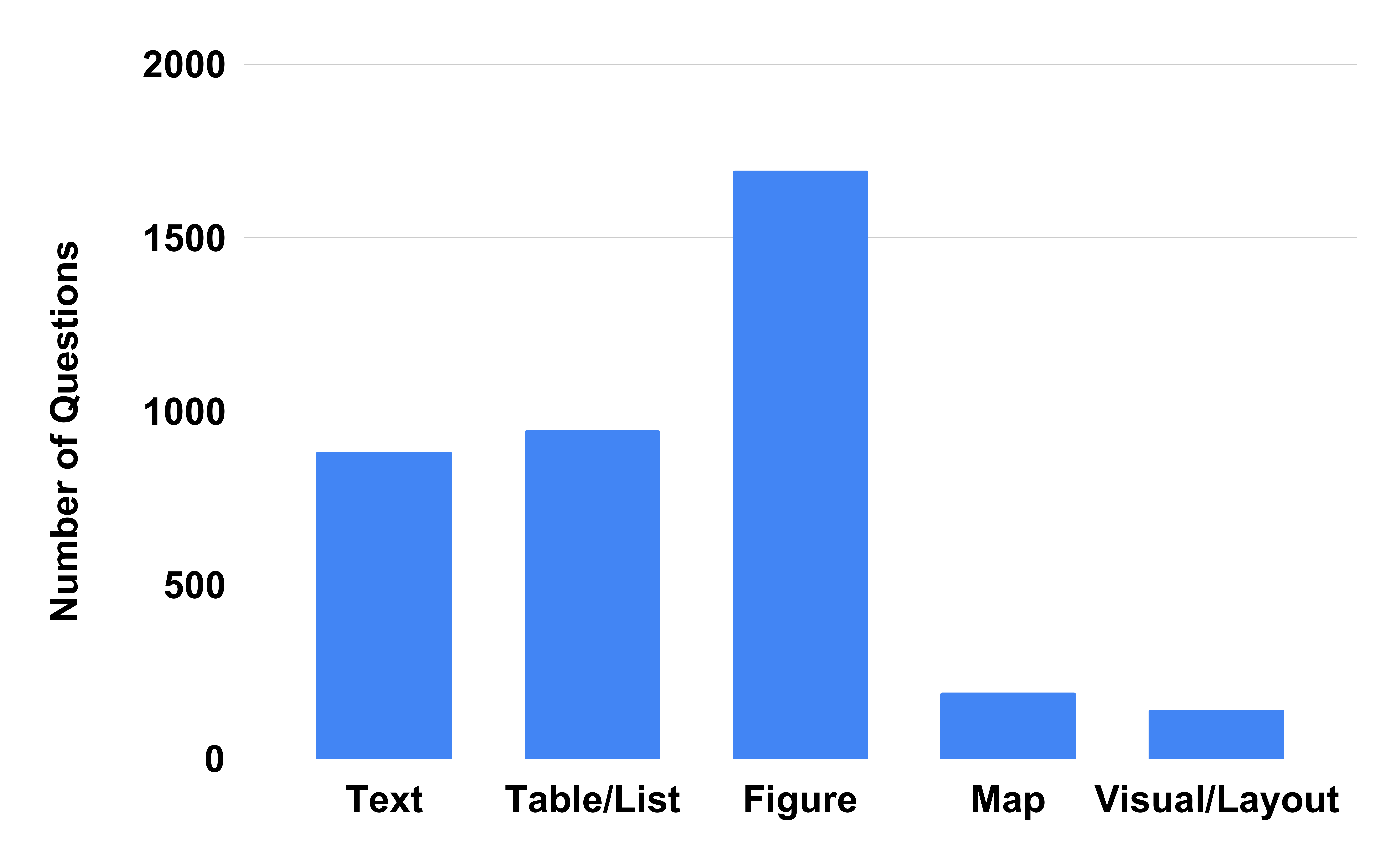}
        \vspace{-5mm}
        \caption{Evidence types and their counts}
        \label{fig:evidence_types}
    \end{subfigure}
    \hspace{1mm}
    \begin{subfigure}[t]{0.32\textwidth}
        \includegraphics[width=1\linewidth]{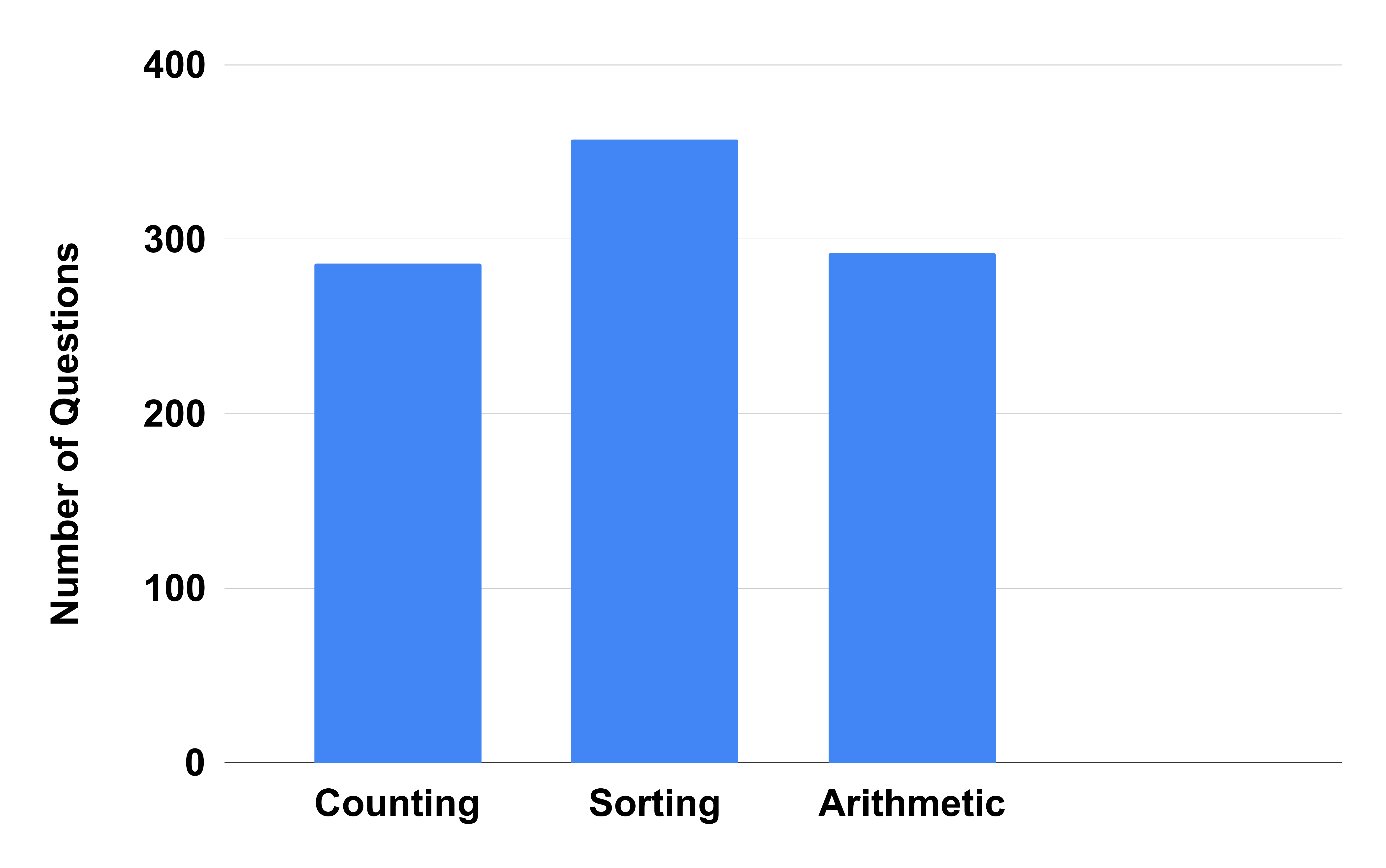}
        \vspace{-5mm}
        \caption{Operation types and their counts}
        \label{fig:operation_types}
    \end{subfigure}
    \vspace{-2mm}
    \caption{Count of questions in validation set by their Answer-source, (\ref{fig:answer_types}), Evidence required to answer (\ref{fig:evidence_types}) and  the discrete Operation performed to find the answer (\ref{fig:operation_types}).}
    \label{fig:type_counts}

\end{figure*}

The success of transformer-based models for text understanding  inspired the use of similar models for document iamge understanding.  \layoutlm and  \lambert~\cite{lambert} incorporate layout information to the BERT architecture by using embeddings of the 2D positions of the text tokens in the image. One of the strong baselines we use in this work is based on the \layoutlm model.
Concurrent to this work, there have been multiple works published on arXiv that deal with joint understanding of text, image and layout in document images. Models such as LayoutLMv2~\cite{layoutlm2}, TILT~\cite{full_tilt_boogie}, DocFormer~\cite{docformer} and StrucText~\cite{structext} build on transformer-based architectures and leverage large-scale pretraining on unlabelled data, using pretraining objectives specifically designed for document understanding.

\textbf{Infographics understanding.} %
Bylinskii \etal~\cite{infographics_tag} and Madan \etal~\cite{icon_detection_infographic} looked at generating textual and visual tags from infographics.
In another work, Landman uses an existing text summarization model to generate captions for infographics~\cite{infographic_captioning}. This model uses only text recognized from infographics to generate the captions and layout/visual information is not considered. These three works use \visuallydataset dataset comprising images from a single website.
MASSVIS~\cite{massvis} is a collection of infographics created to study infographics from a cognitive perspective. As observed by Lu \etal ~\cite{vif_Lu}, MASSVIS is a specialized collection focusing on illustration of scientific procedures and statistical charts, therefore not representative of general infographics. %

To summarize, existing  datasets containing infographics
are either specialized collections  or infographics collected from a single source. 
In contrast, the \datasetname dataset comprises infographics drawn from thousands of different sources, with diverse layouts and designs, and without any topic specialization.

\vspace{-3mm}
\section{\datasetname}
\label{sec:dataset}

A brief description of the data collection and detailed analysis of the data is presented here.
For more details on data collection, annotation process and annotation tool, refer to Section A in the supplementary material.
\vspace{-1mm}
\subsection{Collecting images and question-answer pairs}
\label{subsec: questions_answers}

Infographics in the dataset were downloaded from the Internet for the search query ``infographics'' . The downloaded images are cleaned for removal of duplicates before adding them to the annotation tool.
Unlike crowd sourced annotation, \datasetname was annotated by a small number of annotators using an internal annotation tool. 
The annotation process involved two stages. In the first stage, workers were required to add question-answer pairs on an image shown. 
Similarly to SQuAD dataset~\cite{squad} annotation, in order to make the evaluation more robust, we collect an additional answer for each question in the validation and test split by sending those questions through a second stage of annotation. In this stage an image along with questions asked on it in the first stage are shown to  a worker. Workers were instructed to enter answers  for the questions shown or flag a question if it is unanswerable.

\vspace{-1mm}
\subsection{Question-answer types: answer-source, evidence and operation }
\label{subsec:types_evidence_operation}
In the second stage of annotation, in addition to answering questions collected in the first stage, we instructed the workers to add question-answer types (QA types). QA types are a set of  category labels  assigned to each question-answer pair.
\docvqa and \visualmrc  have QA types that indicate the kind of document object (table, form, title etc.)  a question is based on.
DROP~\cite{drop} dataset for  reading comprehension define answer types such as Question span and Passage span and  categorize questions by the kind of discrete operations ( count, add etc.)  required to find the answer. 
In \datasetname we collect QA types under three categories --- Answer-source, Evidence and Operation.

There are four types of Answer-source --- Image-span, Question-span, Multi-span and Non-extractive. Akin to the definition of `span' in SQuAD~\cite{squad} and \docvqa, an  answer is  considered Image-span if it corresponds to a single span (a sequence of text tokens) of text, extracted verbatim, in the reading order, from text present in the image. Similarly when the answer is a span from the question it is labelled as Question-span. In~\autoref{fig:intro_fig} the answer to the second question is both an Image-span and a Question-span since `canada post' appears as a single sequence of contiguous tokens (or a `span')  both in the question and in the image.
A Multi-span answer is composed of multiple spans of text from the image.
Like in the case of \drop dataset annotation, we instructed our workers to enter Multi-span answers by separating each individual span  by a comma and a white space. For example in \autoref{fig:intro_fig} for the last question the answer is names of two cities. And the names do not appear in a contiguous sequence of text.
Hence it is a Multi-span answer. For a Multi-span answer, any order of the individual spans is a valid answer. In case of the above example, both  ``Vancouver, Montreal'' and ``Montreal, Vancouver'' are valid answers. Since such answers are unordered lists, at evaluation time we consider all permutations of the list as valid answers for the question. The 'Non-extractive' type is assigned when the answer is not an extracted one. While collecting question-answer pairs, Non-extractive questions were allowed only if the answer is a numerical value.
Inclusion of Question-span, Multi-span and numerical Non-extractive answers in \datasetname is inspired by similar setting in the \drop dataset. 
We see this as a natural next step in VQA involving text, different from the purely extractive QA setting in datasets like \docvqa and \stvqa, and abstractive question answering in \visualmrc where automated evaluation is difficult. 
By allowing only numerical answers in Non-extractive case, we make sure that such answers are short and unique, giving no room for variability. 
Near perfect human performance while using automatic evaluation metrics (\autoref{table:_heuristics}) validates that answers in \datasetname are unique with minimal variability when answered by different individuals.

The Evidence type indicates the kind of evidence behind the answer. Types of evidence are Text, Figure, Table/List, Map and Visual/Layout. For example Map is used if the question is based on data shown on a geographical map. Visual/Layout type is added when evidence is based on the visual or layout aspect of the image. For example, questions such as ``What is the color of the hat - brown or black?'' or ``What is written at the top left corner'' fall in this category. Sometimes it is difficult to  discern evidence for a question-answer pair. For example, for the first question in~\autoref{fig:intro_fig}, although the evidence type assigned by the worker is `Figure', it could even be `Table/List' since the visualization looks like a table.
The operation type captures the kind of discrete operation(s)  required to arrive at an answer --- Counting, Arithmetic or Sorting.

In~\autoref{fig:type_counts} we show the distribution of questions in the validation split based on Answer-source, Evidence and Operation. As evident from~\autoref{fig:intro_fig}, a question can have multiple types of answer source, evidence, or operation and many questions do not require any of the specified discrete operations to find the answer. For these reasons, counts in plots shown in~\autoref{fig:type_counts} do not add up to $100\%$.

\begin{figure}[b]
    \centering
    \includegraphics[width=0.8\linewidth]{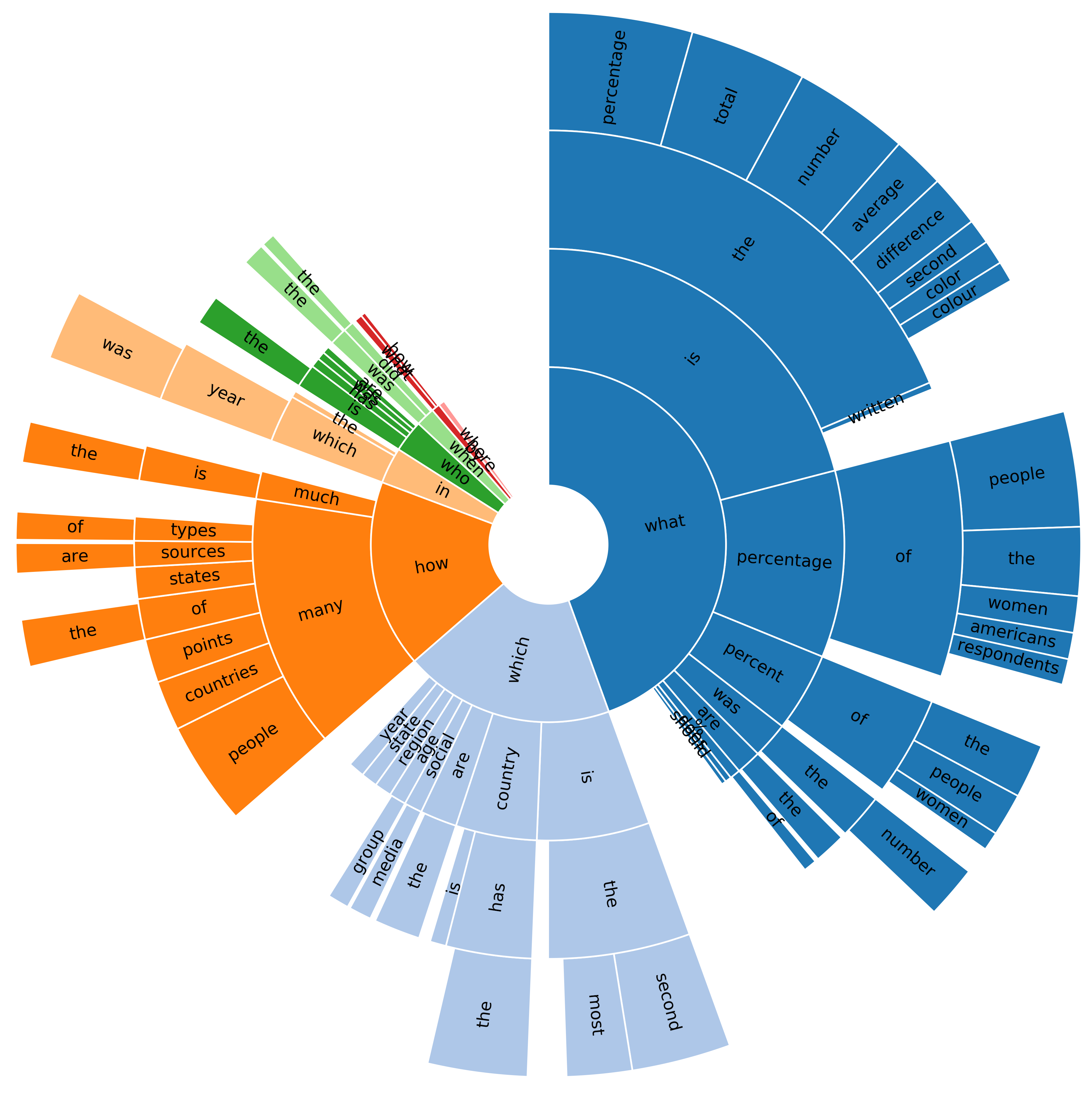}
    \vspace{-3mm}
    \caption{Staring 4-grams of common questions in \datasetname.}
    \label{fig:sunburst_4grams}
\end{figure}

\vspace{-1mm}
\subsection{Summary, statistics, and analysis}
\label{subsec:stats_analysis}
\begin{figure}[h]
    \centering
    \includegraphics[width=0.48\linewidth]{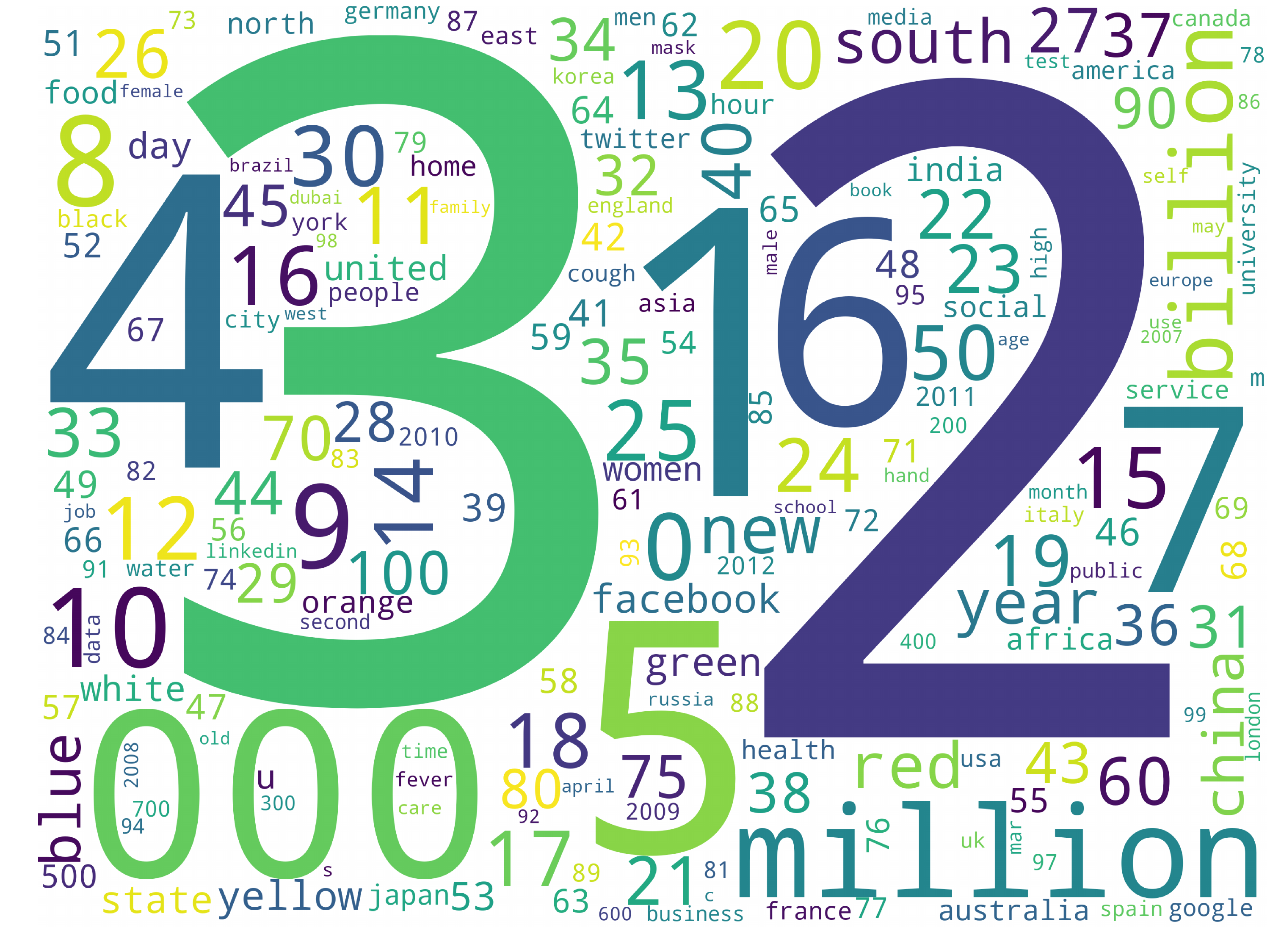}
    \includegraphics[width=0.48\linewidth]{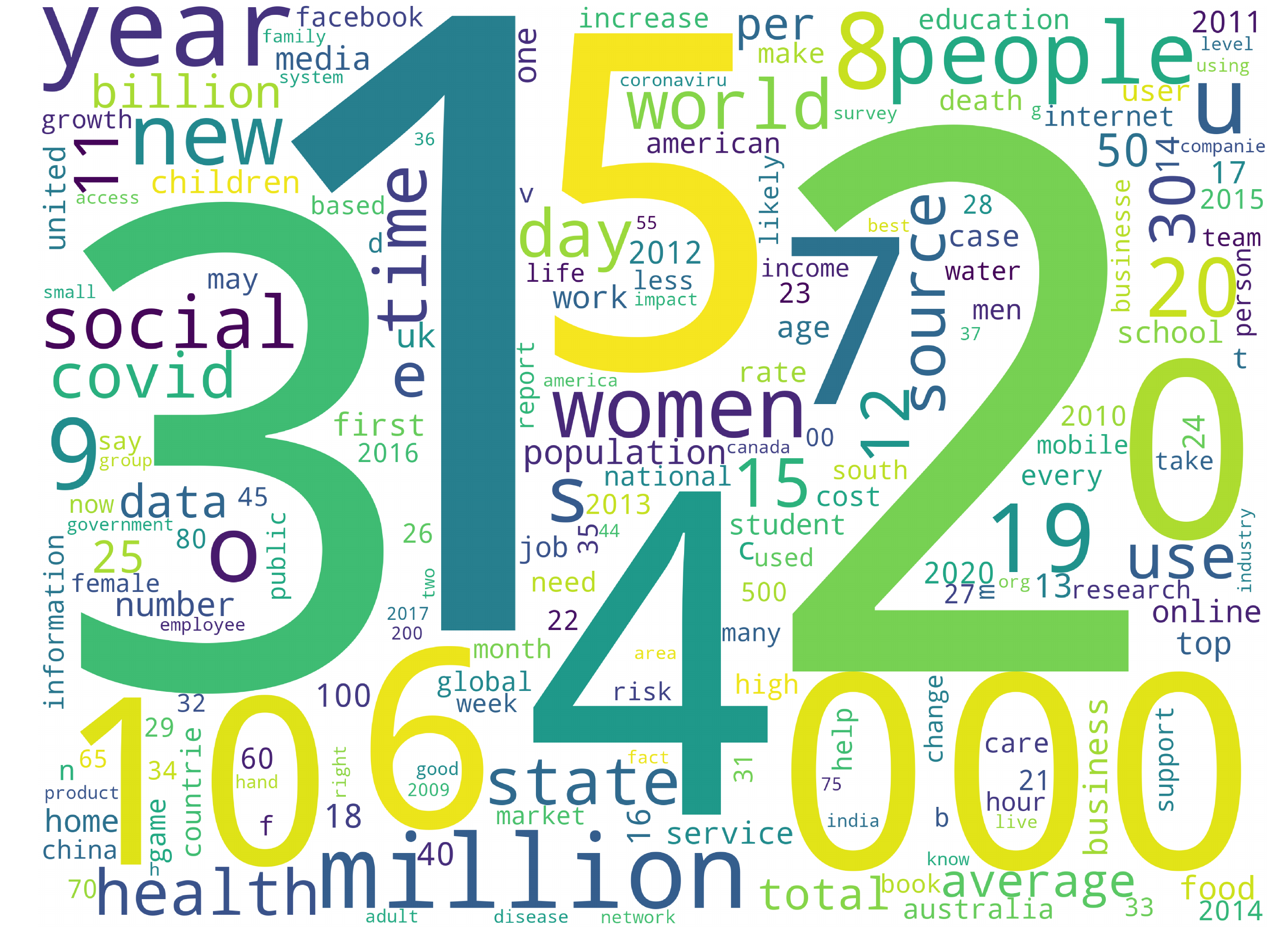}
    \vspace{-3mm}
    \caption{Word cloud of answers (left) and word cloud of words recognized from the infographics (right).}
    
    \label{fig:wordcloud}
\end{figure}
\setlength{\tabcolsep}{2pt}  %

\begin{table}[ht]
\footnotesize
\begin{center}
\begin{tabular}{lrrrrrrr}
\toprule
\multirow{2}{*}{Dataset} & \multicolumn{2}{c}{Questions} &  & \multicolumn{2}{c}{Answers} &  & Avg. tokens   \\
                         & \%Unique      & Avg. len     &  & \%Unique     & Avg. len    &  & per image   \\
\midrule
\stvqa       & 84.84    & 8.80      & & 65.63     & 1.56      & &   7.52              \\
\textvqa     & 80.36    & 8.12      & & 51.74     & 1.51      & &  12.17              \\
\visualmrc   & 96.26    & 10.55     & & 91.82     & 9.55      & & 151.46              \\
\docvqa      & 72.34    & 9.49      & & 64.29     & 2.43      & & 182.75              \\
\datasetname & 99.11    & 11.54     & & 48.84     & 1.60      & & 217.89              \\
\bottomrule

\end{tabular}
\vspace{-3mm}
\caption{Statistics of questions, answers and OCR tokens in \datasetname and other similar VQA datasets. }
\label{tab:datasets_que_answ_stats}
\end{center}
\end{table}

\setlength{\tabcolsep}{6pt}  %

\datasetname dataset has $30,035$ questions and $5,485$ images in total. These images are from  $2,594$ distinct web domains. The data is split randomly to $23,946$ questions and $4,406$ images in train, $2,801$ questions and $500$ images in validation, and $3,288$ questions and $579$ images in test splits. We show basic statistics of questions, answers and OCR tokens in \datasetname and other similar datasets in~\autoref{tab:datasets_que_answ_stats}.

\noindent\textbf{Questions.} ~\autoref{tab:datasets_que_answ_stats} shows that \datasetname has highest percentage of unique questions and highest average question length compared to other similar datasets. \autoref{fig:sunburst_4grams} shows a sunburst of the common questions in the dataset.
There are a  good number of  questions asking for ``How many...'' or percentages. This is expected since infographics carry a lot of numerical data. %
\newcommand{\questionstyle}[1]{\footnotesize \textcolor{black}{\texttt{\textbf{#1}}}}
\begin{figure*}[t]
\begin{center}

\begin{subfigure}[t]{0.25\textwidth}
    \includegraphics[width=\linewidth]{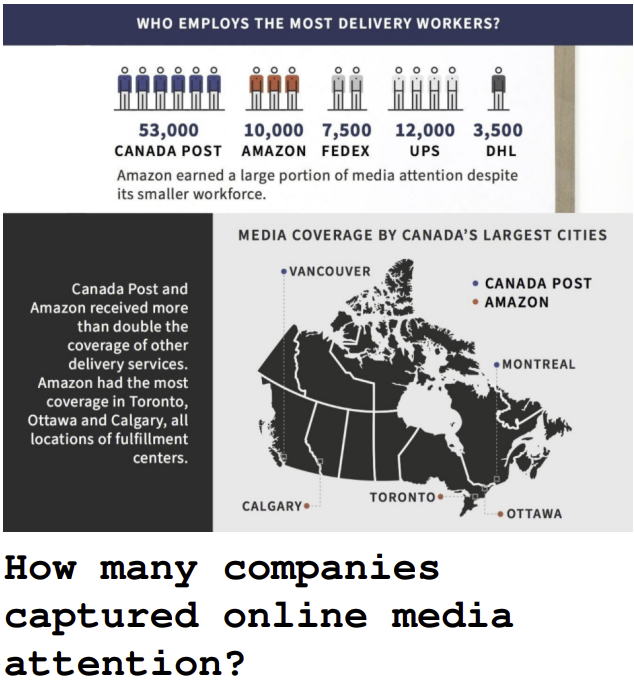}

\end{subfigure}
\hspace{1mm}
\begin{subfigure}[t]{0.70\textwidth}
    \includegraphics[width=\linewidth,height=0.4\linewidth]{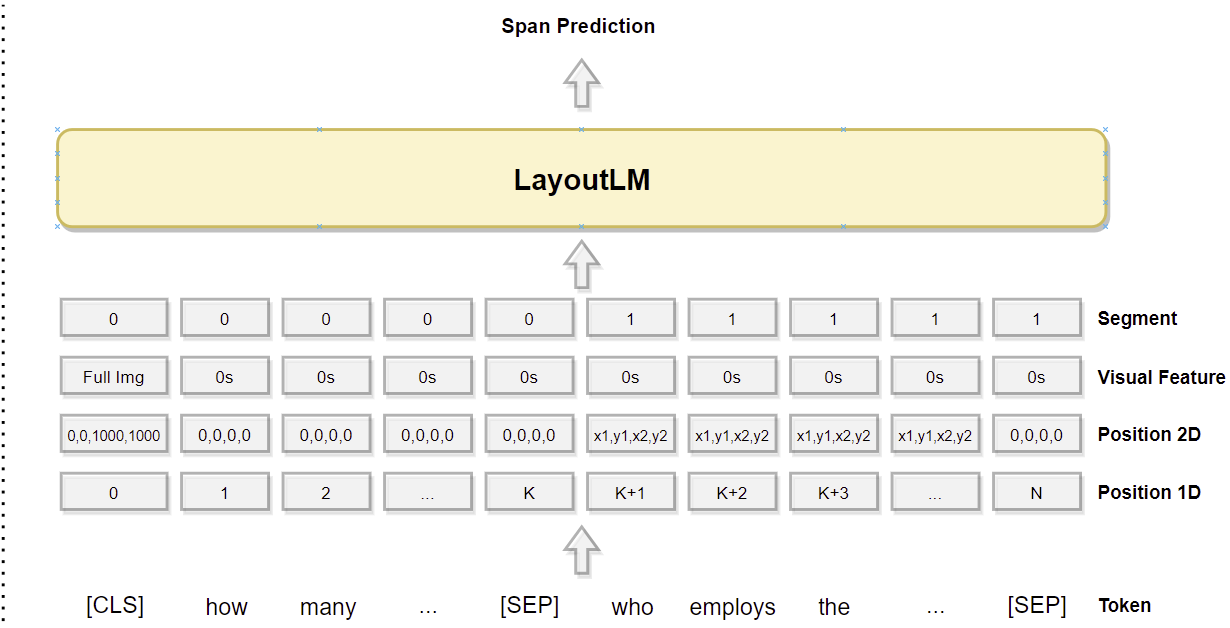}
    
\end{subfigure}
\end{center}
\vspace{-3mm}
  \caption{Overview of our LayoutLM based model for predicting answer spans. Textual, visual and layout  modalities are embedded and mapped to the same space. Then, they are added and passed as input to a stack of Transformer layers.}
\label{fig:model}
\end{figure*}

\noindent \textbf{Answers.}  %
It can be seen in~\autoref{fig:wordcloud} that the most common answers are numbers. This is the reason why \datasetname has a smaller number of unique answers and  smaller average answer lengths.

\noindent \textbf{Embedded text.} The number of average text tokens in \datasetname images is $217.89$. 
Word cloud  of Optical Character Recognition (OCR) tokens spotted on the images is shown in the wordcloud on the right in~\autoref{fig:wordcloud}. Common answers and common text tokens in images are similar.

\vspace{-2mm}
\section{Baselines}
\label{sec:baselines}

In this section, we describe the baselines  we evaluated on the \datasetname. These include heuristic baselines and upper bounds, and SoTA models for VQA and document understanding.

\vspace{-1mm}
\subsection{Heuristic baselines and upper bounds}
\label{subsec:heuristics_upperbounds}

Heuristic baselines and upper bounds we evaluate are similar to the ones evaluated in other VQA benchmarks like \textvqa and \docvqa.

\noindent \textbf{Heuristic Baselines.}
We evaluate performance when answer predicted for each questions is i) \textbf{Random answer} from the train split, ii) \textbf{Majority answer} from the train split and iii) \textbf{Random OCR token} from the image on which question is asked.

\noindent \textbf{ Upper bounds.} We evaluate the performance upper bound on predicting the correct answer if the answer is present in a vocabulary of most common answers in train split. This upper bound is called \textbf{Vocab UB}.
Following \docvqa, to assess the percentage of questions for which answers can be found from the given OCR transcriptions, we compute \textbf{OCR UB} which measures the upper bound on performance if we always predict the correct answer, provided answer is a sub sequence of the serialized OCR transcription of the given infographic. We serialize OCR tokens in the natural reading order , i.e., from top-left to bottom-right. \textbf{Vocab+OCR UB} is the percentage of questions which are either in Vocab UB or OCR UB.

\vspace{-1mm}
\subsection{M4C}
\label{subsec_m4c}
M4C uses a  Transformer stack to fuse representations of a question, OCR tokens, and image. Answers are predicted using an iterative, auto-regressive decoder decoding one word at a time, either from a fixed vocabulary or from the OCR tokens spotted on the image.
Original M4C uses Region of Interest (ROI) pooled features from Box head of a Faster-RCNN~\cite{faster-r-cnn} as the bottom-up visual features. 
Visual features are used for both the objects detected on the image and the OCR tokens spotted on the image. 

\vspace{-1mm}
\subsection{LayoutLM}
\label{sec:method}

LayoutLM~\cite{layoutlm1} extends BERT by incorporating layout information into the original BERT model. LayoutLM pretrained on millions of document images has proven to be effective for multiple document image understanding tasks.
We adapt LayoutLM for InfographicVQA by changing the input to suit a multimodal setting and using an output head for SQuAD-style span prediction.
Since we are using SQuAD-style span prediction at the output, this model can only handle questions whose Answer-source is Image-span. Nearly $70\%$ of questions in validation and test split are of this type. Extending this model to include  questions with other Answer-sources is an interesting direction for future work.
\vspace{-1mm}
\subsubsection{Model overview}
A schematic of the LayoutLM-based model which we use for \datasetname is shown in~\autoref{fig:model}.
Input sequence of tokens to the model are composed of  the tokens in question and the  tokens spotted on the image by an OCR.
The sequence starts with a special [CLS] token, followed by the question tokens and the OCR tokens. The sequence ends with a special [SEP] token. Question and OCR tokens are also separated by a [SEP] token. All the tokens in the input sequence are represented by a corresponding embedding which in turn is the sum of i) a token embedding, ii) a segment embedding,  iii) a 1D position embedding, iv) Four 2D position embeddings, and v) a visual embedding.

Following the original setting in BERT, for  \textbf{token embedding,} we use WordPiece embeddings~\cite{wordpiece} with a $30,000$ size vocabulary. These embeddings are of size $768$ . A  \textbf{segment embedding} differentiates different segments in the input. For example when the input sequence is formed by tokens from question and OCR tokens, question tokens and OCR tokens  are given a segment id of $0$  and $1$ respectively.
\textbf{1D position embedding} is used to indicate the order of a token within the sequence. For OCR tokens we use the default reading order by serializing the tokens in top-left to bottom-right order.

For 2D position embedding, we follow the same process as original  LayoutLM. Given the bounding box coordinates $(x_1,y_1,x_2,y_2)$ of an OCR token, we embed all the 4 coordinate values using 4 separate embedding layers. $x_1$ and $x_2$ share same embedding table and $y_1$ and $y_2$ share another common embedding table. The coordinate values are normalized to lie in the range 0--1000 before embedding. For [CLS] token we use 2D embedding corresponding to $(0,0,1000,1000)$. For question tokens all the four 2D position embeddings used are 0s.

Unlike in original LayoutLM where visual features are fused  after getting the attended embeddings from the Transformer block, we fuse the visual features early with the text input. 
Similar to M4C, for each OCR token, we use ROI pooled feature from the Box head of a pretrained object detection style model. This
feature is mapped to same size as other embeddings using a linear projection layer.
For [CLS] and other special tokens we add visual feature corresponding to an ROI covering the entire image, named as ``Full Img'' in~\autoref{fig:model}.

\vspace{-1mm}
\subsubsection{Training procedure}

Similar to the BERT and  original LayoutLM, we trai the model in two stages.

\noindent \textbf{Pretraining}: Following original LayoutLM, we use Masked Visual-Language Model (MVLM) task for pretraining with a masking probability of $0.15$.
Whenever masking, we replace each token with the [MASK] token 80\% of the time, with a random token 10\% of the time and keep it unchanged 10\% of the time. 

\noindent\textbf{Finetuning}: For finetuning, similar to BERT QA model for SQuAD benchmark, the output head used in pretraining is replaced by a span prediction head that predict start and end token positions of the answer span.

\vspace{-3mm}
\section{Experiments and results}
\label{sec:experiments}
In this section, we describe the experiments  we conducted on the new \datasetname dataset and present the results and analysis.
\setlength{\tabcolsep}{3pt}  %

\begin{table}[b]
\footnotesize
\begin{tabularx}{\linewidth}{lcccccc}
\toprule
Detector & \multicolumn{2}{c}{\textvqa} & \multicolumn{2}{c}{\docvqa} & \multicolumn{2}{c}{\datasetname} \\
& Avg.  &  \textless 2 det.(\%) & Avg. &   \textless 2 det.(\%) & Avg.  &  \textless 2 det.(\%) \\
\midrule

VG & 28.8  & 0.0 & 4.1 & 43.9 & 7.4 & 23.9 \\
DLA & 1.0 & 97.9  & 4.7 & 0.0 & 2.9 & 43.4 \\

\bottomrule
\end{tabularx}
\vspace{-3mm}
\caption{\small Statistics of object detections using two detectors -- VG and DLA. DLA is trained for document layout analysis and VG is an object detection model trained on Visual Genome. Avg. shows average number of detections per image. ` \textless 2 det.(\%)' is the percentage of images on which number of detected objects is less than 2. Results show that DLA and VG which are suitable for document images and natural images respectively, detect much lesser object instances on infographics.}
\label{tab:obj_detection_stats}

\end{table}

\vspace{-1mm}
\subsection{Experimental setup}
\label{subsec:exp_Setup}

\noindent \textbf{Evaluation metrics. }
For evaluating VQA performance on \datasetname, we use Average Normalized Levenshtein Similarity (ANLS) and Accuracy metrics. The evaluation setup is exactly same as evaluation in \docvqa.

\noindent \textbf{OCR transcription.}
Text transcriptions and bounding boxes for text tokens in the images are obtained using Textract OCR~\cite{textract}.

\noindent \textbf{Human performance}
For evaluating human performance, all questions in the test split of the dataset are answered with the help of two volunteers (each question answered by a single volunteer).

\noindent \textbf{Vocabulary of most common answers.}
For Vocab UB and heuristics involving a vocabulary, we use a vocabulary of 5,000 most common answers in the train split.

\begin{table}[t]
\footnotesize
\begin{tabularx}{\linewidth}{X c c c c}
\toprule
&  \multicolumn{2}{c}{ANLS} & \multicolumn{2}{c}{Accuracy(\%)} \\             
  Baseline   & val & test.  & val & test  \\
\midrule

Human performance & - & 0.980 & - & 95.70 \\ 
Random answer   & 0.006 & 0.005 & 0.00 & 0.00 \\  %
Random OCR token   & 0.011 & 0.014 & 0.29 & 0.49 \\  %
Majority answer  & 0.041 & 0.035 & 2.21 & 1.73 \\  %

Vocab UB   &-  & - & 53.16 & 51.34 \\
OCR  UB &-  &  - & 53.95 & 56.96 \\
Vocab + OCR UB & - & -  & 76.71 & 77.4 \\
\bottomrule
\end{tabularx}
\vspace{-3mm}
\caption{\small Results of heuristics and upper bounds.  Heuristics yield near zero results. More than $75\%$ of the questions have their answer present either in a fixed vocabulary or as an Image-span of the OCR tokens serialized in default reading order. }
\label{table:_heuristics}

\end{table}

\noindent \textbf{ROI Features.}
For our experiments using M4C and LayoutLM models, visual features of different bounding regions from the images are used. To this end, we use two pretrained object detection models --- a Faster-RCNN~\cite{faster-r-cnn} trained on Visual Genome~\cite{visual_genome} and a Mask-RCNN~\cite{maskrcnn} trained on document images in PubLayNet~\cite{publaynet} for Document Layout Analysis (DLA). We refer to these detectors as VG and DLA respectively in our further discussion.  The FasterRCNN model we use is same as the one used for M4C. 
We use the implementation in MMF framework~\cite{mmf}. 
The DLA detector we use is from a publicly available Detectron2~\cite{wu2019detectron2}-based implementation~\cite{dla_detectron2}. Features from the last or second last Fully Connected (FC) layer are used as visual features in M4C and LayoutLM model. In VG and DLA, these features are of size $2048$ and $1024$ respectively.

In ~\autoref{tab:obj_detection_stats} we give a summary of the detections using the two detectors on \textvqa, \docvqa and \datasetname. 
With DLA, we notice that many of its detections, especially when there is only one detection per image is a box covering the entire image.

\noindent\textbf{Experimental setting for M4C.} We use the official implementation of the model~\cite{mmf}. The training parameters and other implementation details are same as the ones used in the original paper.
As done in original M4C, fixed vocabulary used with the model is created from $5,000$ most common words among words from answers in the train split.

\noindent\textbf{Experimental setting for \layoutlm.} The  model is implemented in Pytorch~\cite{pytorch}. 
In all our experiments we start from a pretrained checkpoint of LayoutLM model made available by the authors in Hugginface's Transformers model zoo~\cite{huggingface_Arxiv,huggingface_model_zoo}. The newly introduced linear projection layer which maps the ROI pooled features  to the common embedding size of $768$, is initialized from scratch. The features are from the last FC layer of the Box head of DLA or VG.
To continue pretraining using  in-domain data, we use four samples in one batch and Adam optimizer with learning rate $2e-5$. For finetuning, we use a batch size of $8$ and Adam optimizer with learning rate $1e-5$.
For in-domain pretraining and finetuning no additional data other than train split of \datasetname is used.
To map answers in \datasetname train split to SQUAD~\cite{squad}-style spans, we follow the same approach used by Mathew \etal in \docvqa.
We take  the first subsequence match of an answer in the serialized transcription as the corresponding span.
This way we found approximate spans for $52\%$ of questions in the train split. Rest of the questions are not used for finetuning the model.

\vspace{-1mm}
\subsection{Results}
\label{subsec:results}

Results of heuristic baselines, upper bounds, and human performance is shown in~\autoref{table:_heuristics}. Human performance is comparable to the human performance on \docvqa. %
As given by the Vocab + OCR UB, more than 3/4 of questions have their answers present as a span of the OCR tokens serialized in the default reading order or in a vocabulary of most common   answers in the train split.

\begin{table}[t]
\footnotesize
\begin{tabularx}{\linewidth}{X c l c  c c c c}
\toprule
Visual&  Finetune &Object\& & {\#  OCR} &      \multicolumn{2}{c}{ANLS} & \multicolumn{2}{c}{Accuracy(\%)} \\  
{Feature} & detector &  {Count} & {tokens } & val & test & val & test \\

\midrule

VG  & \cmark  &  Obj. (100)& 50 & 0.107 & 0.119 & 4.81 & 4.87 \\
VG  & \cmark &  Obj. (20) & 50 & 0.111 & 0.122 & 4.82 & 4.87 \\
VG   & \xmark &  Obj. (20) & 50 & 0.125 & 0.127 & 4.89 & 4.89 \\
VG   & \xmark &  Obj. (20) & 300 & 0.128 & 0.134 & 4.90 & 5.08 \\
VG   & \xmark &  None   & 300 & 0.136 & 0.143 & 5.86 & 6.58 \\
VG   & \xmark &  Full Img   & 300 & \textbf{0.142} & \textbf{0.147} & 5.93 & 6.64 \\
DLA  & \xmark  &  Obj. (20) & 50 & 0.110 & 0.130 & 4.86 & 5.02 \\
DLA  & \xmark &  Obj. (20) & 300 & 0.132 & 0.144 & 5.95 & 6.50 \\
DLA  & \xmark  &  None   & 300 & 0.140 & 0.142 & 5.90 & 6.39 \\
DLA  & \xmark &  Full Img   & 300 & 0.138 & 0.140 & 5.97 & 6.42 \\

\bottomrule
\end{tabularx}
\vspace{-3mm}
\caption{\small Performance of different variants of the M4C model. The original M4C setting is the one shown in the first row. `Finetune detector' denotes the case when features from penultimate FC layer is used and last FC layer is finetuned along with the M4C model. This is the default setting in M4C. In our experiments, we get better results without finetuning. `Obj. (100)' is the case when features from up to 100  objects (bottom-up features)  are used. We experiment with 20 objects per image and the results did not change much. Using no object (`None') and feature from only one object---a box covering the entire image (`Full Img')---yield better results than the case where bottom-up objects are used.}
\label{tab:m4c_results_singlecolumn}

\end{table}

We show results using M4C model in~\autoref{tab:m4c_results_singlecolumn}. In contrast to the original setting for which finetuning of visual features and  features of detected objects are used, a setting that uses no finetuning and only a single visual feature corresponding to ROI covering the entire image, yields the best result.

\setlength{\tabcolsep}{3pt}  %
\newcommand{\cls}{CLS}
\newcommand{\nonocr}{NORC}
\newcommand{\all}{All}
\newcommand{\que}{Que}
\newcommand{\sep}{SEP}

\newcommand{\dla}{DLA}
\newcommand{\vg}{VG}

\begin{table}[b]
\footnotesize
\begin{center}
\begin{tabularx}{\linewidth}{Xccccccc}
\toprule
Full Img & Visual & Continue & OCR  & \multicolumn{2}{c}{ANLS} & \multicolumn{2}{c}{Accuracy (\%)}  \\
to & feature & pretrain.  & visual& val & test & val & test \\
\midrule
 -    & -               & \xmark   & \xmark     & 0.212  & 0.225 & 13.40 & 15.32 \\
 -    & -               & \cmark   & \xmark     & 0.250  & \textbf{0.272} & 18.14 & 19.74 \\
 \cls & \dla            & \cmark   & \xmark     & \textbf{0.256}  & 0.261 & 18.56 & 19.16 \\
 \all & \dla            & \cmark  & \xmark     & 0.248  & 0.266 & 17.82 & 18.77 \\       
Non-OCR & \dla & \cmark & \cmark     & 0.245  & 0.263 & 17.21 & 18.37 \\
 \cls & \vg             & \cmark  & \xmark     & 0.229  & 0.235 & 16.47 & 16.51 \\
 \all & \vg             & \cmark  & \xmark     & 0.109  & 0.106 &  5.43 &  4.96 \\       
 Non-OCR & \vg & \cmark    & \cmark    & 0.042  & 0.037 &  1.75 &  1.28 \\

\bottomrule

\end{tabularx}
\vspace{-3mm}
\caption{\small Performance of LayoutLM with different input settings. Row 1 and 2 show LayoutLM's performance with and without in-domain pretraining. `Visual Feature' column specify the kind of detector used for visual feature. `OCR visual' indicate whether visual features of the OCR tokens are used or not. `CLS', `All' and `Non-OCR' in `Full Img to' column represent Full Img feature added  only to CLS token, all tokens and  all non OCR tokens respectively.}

\label{tab:layoutlm_results_withbert}
\end{center}
\end{table}

\begin{figure*}[t]
    \centering
    \includegraphics[width=\linewidth,height=5.3cm]{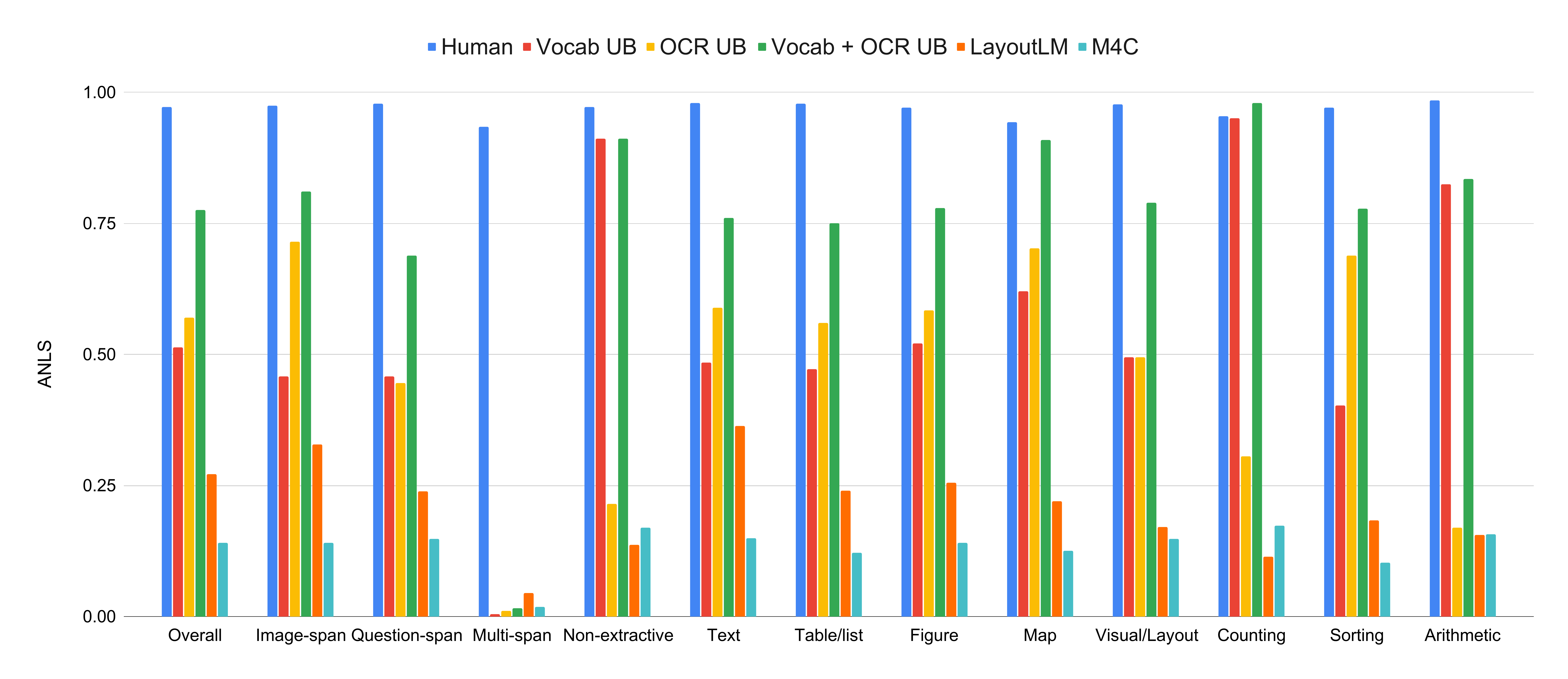}
    \vspace{-6mm}
    \caption{\small Performance of baselines and  upper bounds for different QA types.
    }
    \label{fig:anls_per_type_barchart}
\end{figure*}

\begin{figure*}[h]
\begin{center}
\small
\begin{tabular}{p{0.32\linewidth}p{0.32\linewidth}p{0.32\linewidth}}
    \includegraphics[width=\linewidth,height=0.5\linewidth]{images/10045.jpeg} &
    \includegraphics[width=\linewidth,height=0.5\linewidth]{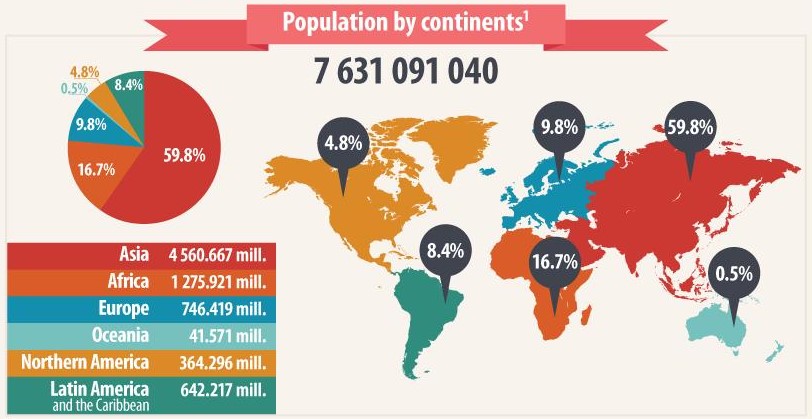} &
    \includegraphics[width=\linewidth,height=0.5\linewidth]{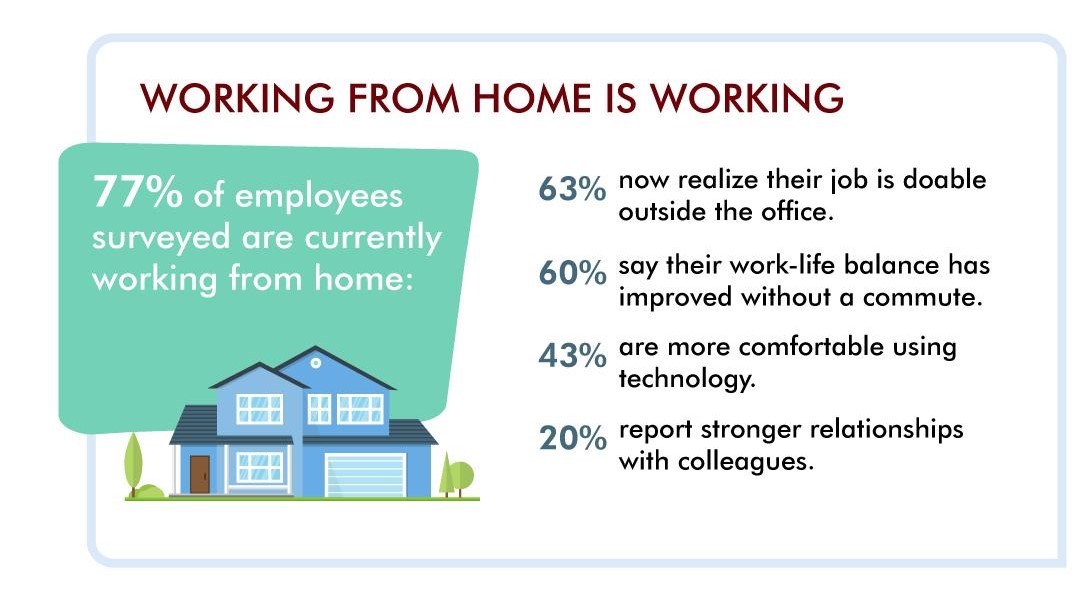} \\
    
    What is the interest rates of European Central Bank and US FED? \par
    LayoutLM: \hspace{0.14cm} \correct{0.25\%} \hspace{0.40cm}
    M4C: \hspace{0.15cm} \wrong{0.1\%} \par
    Human:\hspace{0.66cm} \correct{0.25\%} \hspace{0.40cm}
    GT:\hspace{0.50cm} \groundtruth{0.25\%}
    &
    
    Which is the least populated continent in the world? \par
    LayoutLM: \hspace{0.05cm} \wrong{EU}  \hspace{0.90cm}
    M4C: \hspace{0.10cm} \correct{Oceania} \par
    Human:\hspace{0.60cm} \correct{Oceania} \hspace{0.25cm}
    GT:\hspace{0.40cm} \groundtruth{Oceania}

    &

    What percentage of workers are not working from home? \par
    LayoutLM: \hspace{0.10cm} \wrong{77\%}  \hspace{0.40cm}
    M4C: \hspace{0.15cm} \wrong{66\%} \par
    Human:\hspace{0.60cm} \correct{23\%} \hspace{0.40cm}
    GT:\hspace{0.48cm} \groundtruth{23\%}

\end{tabular}
\end{center}
\vspace{-8mm}
\caption{\small \textbf{Qualitative Results} For the left most question, evidence is Table/List and the \layoutlm model gets the prediction right. In case of the second question where evidence is a Table/List and Sorting is involved, M4C gets the answer correct. In case of the last question which requires subtraction of 77 from 100 neither M4C, nor \layoutlm gets the answer correct. (Since original images are much bigger we only show crops of the infographics which are relevant to the question). More qualitative examples showing images in original size are added in the supplementary.}
\label{fig:qualitative_results}

\end{figure*}

Results of the \layoutlm based model are shown in~\autoref{tab:layoutlm_results_withbert}. In-domain pretraining, using text from question, and OCR tokens helps the model significantly. This is inline with observation by Singh \etal that pretraining 
on data similar to the data for downstream task is highly beneficial in visio-linguistic pretraining~\cite{are_We_pretraining_it_right}.
The model that uses Full Img feature from DLA, added to the CLS performs the best on validation set. And on test set, a model which does not use any visual feature performs the best.

From~\autoref{tab:layoutlm_results_withbert}, it is evident that models which use visual features of OCR tokens do not give better results. 
This implies that token embeddings of the relatively noise free OCR tokens are good enough and the additional information from visual features of the tokens contribute little to the performance.

Most of the recent models that employ visio-linguistic pretraining of  BERT-like architectures~\cite{visualbert,vlbert} incorporate bottom-up visual features---features of objects detected on the images---into the model as visual tokens.
We follow approach in VisualBERT~\cite{visualbert} where   visual tokens are concatenated after  the  input stream of text tokens. Each visual token is represented by a dummy text token [OBJ], a separate segment, 1D and 2D positions and the ROI pooled visual feature of the object's region. But in our experiments, the addition of visual tokens did not give us results any better than the model without visual tokens.
Hence we do not show this setting in illustration of our model architecture or in the results table. We believe the visual tokens we use impart little information since the object detectors we use--- a detector trained for detecting objects on natural scene images and another for document layout analysis---are not suitable for infographics.
This is quite evident in~\autoref{tab:obj_detection_stats}. Both the detectors detect only a few instances of objects on infographics.

In~\autoref{fig:anls_per_type_barchart} performance of  our trained baselines are compared against the upper bounds and human performance, for different Answer-sources, Evidence  and Operation types in test split of the dataset. The M4C and \layoutlm models used for this plot are the variants which give best ANLS on the test data.
Finally a few qualitative results from our experiments are shown in~\autoref{fig:qualitative_results}.

\vspace{-3mm}
\section{Conclusion}
\label{sec:conclusion}
We introduce the \datasetname dataset and associated task of VQA on infographics. Results using the baseline models suggest that existing models designed for multimodal QA or VQA perform poorly on the new dataset.
We believe our work will inspire research towards understanding images with complex interplay of layout, graphical elements and embedded text.

\appendix
\counterwithin{figure}{section}
\counterwithin{table}{section}
\counterwithin{equation}{section}

\newpage

\section{Data collection }
This section provides details on how we collected infographics from internet, cleaned it and the way we collect questions and answers on these images.
\subsection{Collecting images and de-deuping}
Images in the dataset are sourced from internet. Initially we downloaded more than 10K images for the search query ``infographics'' using Google and Bing image search engines. The downloaded images were fist de-duped using using a Perceptual Hashing approach implemented in \textit{imagededup} library~\cite{imagededup} this removed nearly $2000$ duplicate images. The first round of de-duplication helped to reduce the number of images for the second round of de-duplication that involved use of a commercial OCR. In this round, we compared the images using Jaccard similarity of the text tokens spotted on the images using the Amazon Textract OCR~\cite{textract}. After two rounds of de-duplication, around 7K images were left. These were added to the annotation system for question-answer annotation.

\subsection{Question-answer annotation}
Here we present details of the annotation such as selection of workers, annotation process and the annotation tool. \\

\noindent\textbf{Annotation scheme and selection of workers.}

Initially we had hosted  a pilot annotation on a crowd-sourcing  platform for collecting question-answer pairs on infographics. But more than $40\%$ of the question-answer pairs added during the pilot annotation were noisy. 
We realized that some of the requirements were not easy to understand from written instructions. For example  the kind of question-answers which are allowed---only questions whose answer sources are one of the four types---is better understood when explained using examples.
Consequently we decided  to use an internal web-based tool for the annotation and hired workers with whom we could interact closely.

\begin{figure*}[b]
    \centering
    \includegraphics[width=0.8\linewidth]{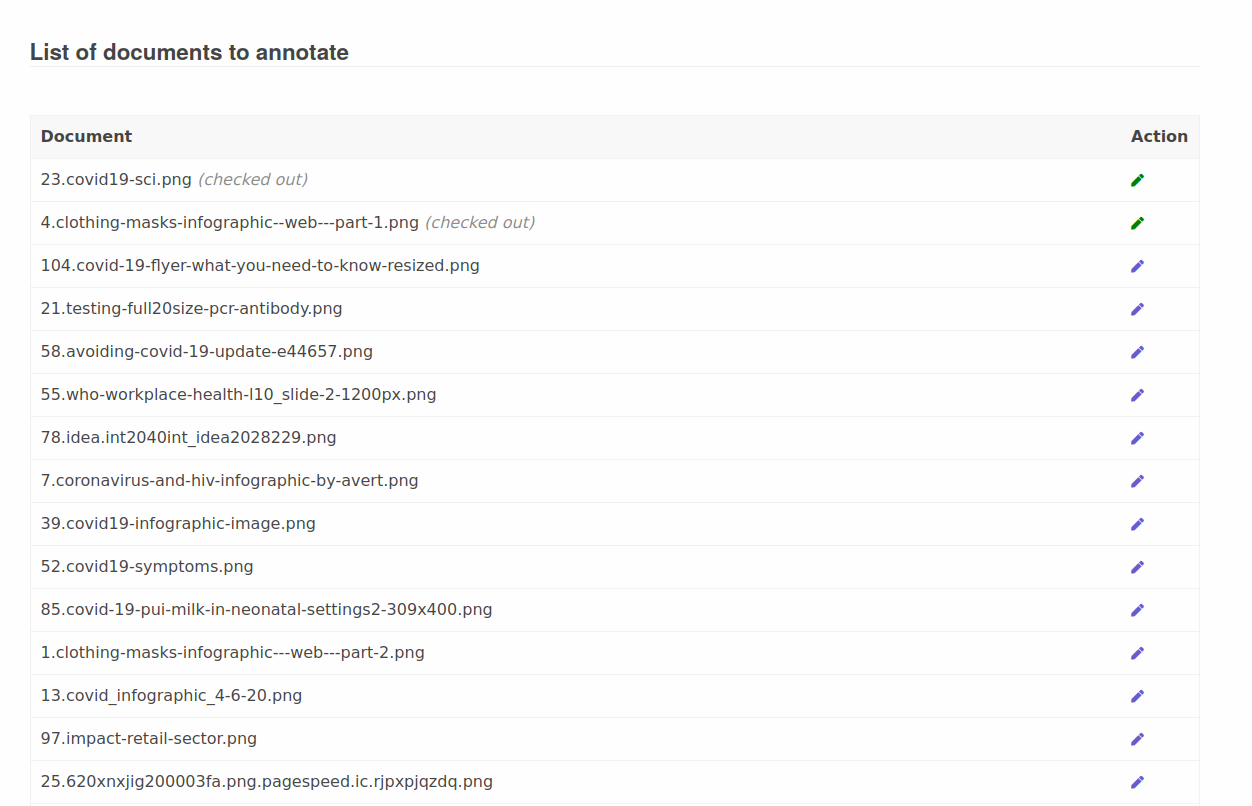}
    \caption{\textbf{Images queue for question-answer collection.} The screenshot shows interface from which a worker picks an image for question-answer collection (stage 1 of annotation). It shows the list of all images in the system that are yet to be annotated, and the images that are already opened (checked out) by the particular user.  In this case  there are two images that are being checked out; the two images shown at the top of the list  Workers were allowed to check out at most 5 images. This feature allows workers to keep documents in pending if they are in doubt. }
    \label{fig:annotation_images_queue}
\end{figure*}

To select the workers, we reached out to individuals looking for annotation-type jobs through mailing lists and other online groups.
Interested applicants were invited to join a 90 minute webinar explaining  the process and all the requirements. During the webinar we explained them each of the instructions with many examples for accepted type of questions and question-answer types. Following the webinar, the applicants were asked to take an online quiz to assess how well they understood the process and the  policies. Base on the quiz scores we selected 13 of them for the annotation. The selected workers were called for another round of webinar where we discussed the answer key for the quiz and clarified their doubts.
The workers were added to an online forum so that they could post their queries related to the annotation in the forum.
They were encouraged to post questions  with screenshots whenever in doubt. They  would keep the particular image in pending and move on to other images in the queue,  until one of the authors give a reply to the question they raised.
 This way we could reduce annotation errors drastically. \autoref{fig:annotation_images_queue} shows a screenshot of our web-based annotation tool that shows the interface from which a worker picks the next image for annotation, while having a few documents in pending.
 \\

\noindent\textbf{Annotation tool}

\begin{figure*}[b]
    \centering
    \includegraphics[width=0.8\linewidth]{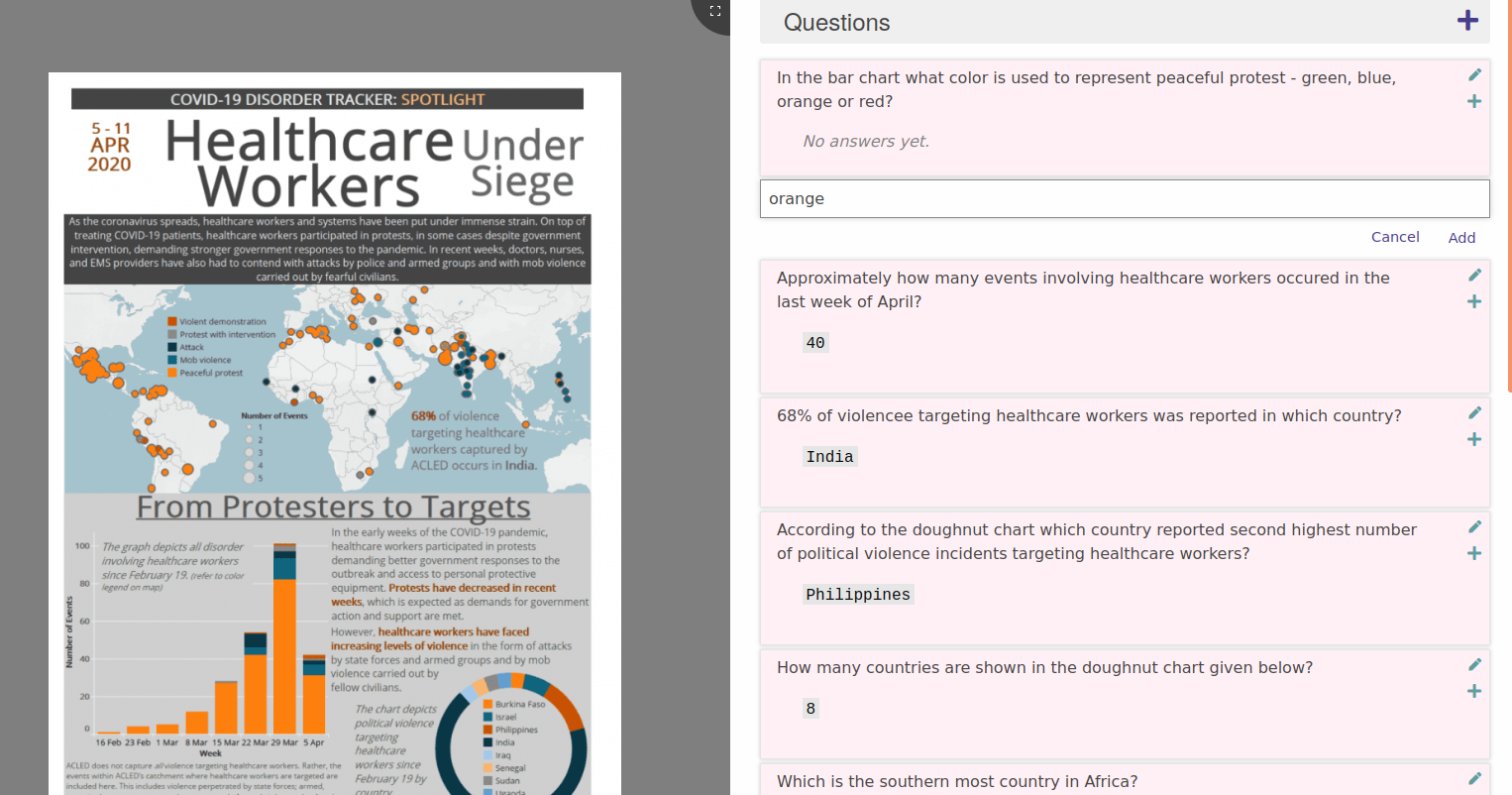}
    \caption{\textbf{First stage of annotation.} Questions and answers are collected at this stage. Image is shown on the left pane with options to zoom, rotate etc. Questions and answers are added on the right pane.} 
    \label{fig:annotation_stage1}
\end{figure*}

Annotation of \datasetname was organized in two stages. In the first stage, questions-answer pairs were collected on the infographics.
Workers were allowed to reject an image if it is not suitable. See~\autoref{table:table_annotation_instructions} for instructions on when to reject an image. After collecting  more than  30K questions and their answers, we stopped the first round of annotation, as we were aiming for a dataset with 30K questions in total. We split this data into train, validation and test splits so that the splits  roughly have 80, 10 and 10 percentage of the total questions respectively. \autoref{fig:annotation_stage1} shows a screenshot from the stage 1 of the annotation.

\begin{figure*}[b]
    \centering
    \includegraphics[width=0.8\linewidth]{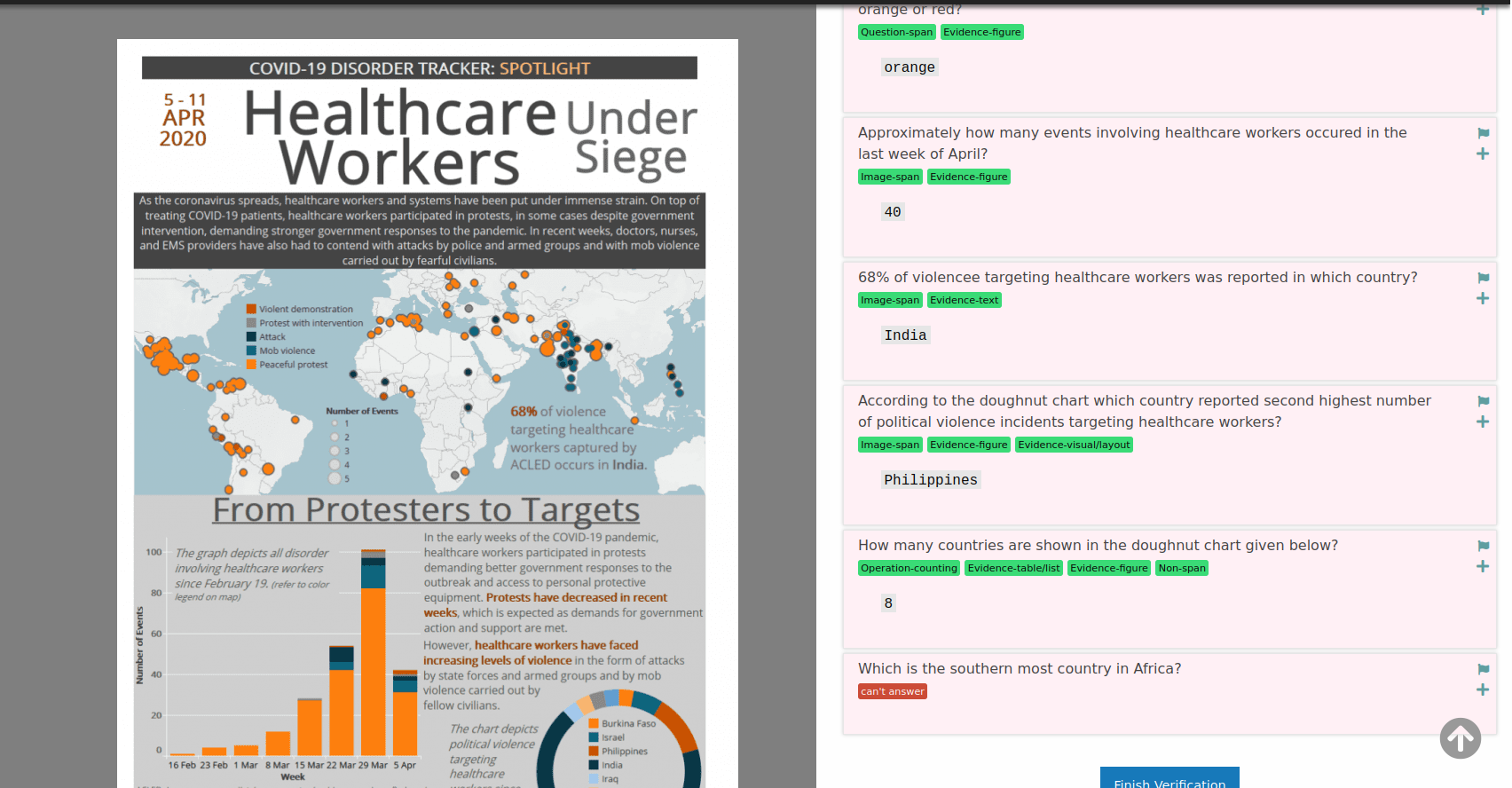}
    \caption{\textbf{Second stage of annotation.} Additional answers and question-answer types  are collected for validation and test splits. For one of the questions, answer cannot be found from the image alone and hence the worker assigned a ``cant answer'' flag. Such questions are removed from the dataset. }
    \label{fig:annotation_stage2}
\end{figure*}

Inspired by the  the SQuAD dataset annotation, we include a second stage in the annotation process to collect additional answers for questions in the validation and test split. Hence only images from validation and test splits were sent through this stage. 
In this stage, a worker was shown an image and were asked to answer the questions that had been  added on the image in the first stage (answers entered in the first stage were not shown). 
Finally, we retain all the unique answers (i.e, unique strings after converting all answers to lower case) entered for a question. Hence a question can have more than 1 valid answer.
We made sure that second stage was done by a worker different from the one who collected questions and answers on the same image in the first stage. 
The workers were also allowed mark a question as "can't answer" if they are not able to find an answer for the question based on the information present in the image. Around $1.9\%$ questions were marked so and those were removed from the dataset. As mentioned in subsection 3.2 in the main paper, in the second stage, workers were also required to assign question-answer types based on answer source, evidence and operation for each question. 
\autoref{fig:annotation_stage2} shows a screenshot from stage 2 of the annotation.

Written instructions for both stages of the annotation that we shared with the workers are given in~\autoref{table:table_annotation_instructions}.

\begin{table*}[b]
\small
\begin{tabularx}{\linewidth}{p{8cm}|p{8cm} }
\toprule
Stage1 1 instructions & Stage 2 instructions \\
\midrule

\begin{enumerate}
    \item You need to  add questions and corresponding answers based on the given image.
    \item Make sure the questions you ask can be answered purely based on the content in the image
    \item Try to minimize questions  based on textual content. Frame questions which require you to connect multiple elements such as text, structured elements like tables, data visualizations and any other visual elements.
    \item On infographics with numerical data, try to frame questions which require one to do basic arithmetic, counting or comparisons
    \item You are allowed to reject the image if,
    \begin{enumerate}
        \item bad resolution images/ illegible text
        \item image is an infographic template with dummy text
        \item content is almost entirely in a non English language
    \end{enumerate}

    \item The text which forms your answer must be
    \begin{enumerate}
        \item found verbatim in the image as a contiguous sequence of tokens in the reading order
        \item found verbatim in the question as a contiguous sequence of tokens
        \item formed by multiple text pieces where each `piece' is found verbatim in the image as contiguous sequence of text tokens. In such a case when you add the answer separate each item by a comma and a white space.
        \item a number such as 2, 2.2, 2/2 etc..
    \end{enumerate}
    
\end{enumerate}
&
\begin{enumerate}
\item You need to enter answer for the questions shown based on the given image
\item if you cannot find answer to the question based on the image, flag the question as "cant answer"

\item For each question add the following question-answer types appropriately. There are three categories of question-answer types - Evidence, Operation, and Answer-source. It is possible that a question can have more than one source of answer, more than one type of operation, or  more than one type of evidence associated with it.
\begin{enumerate}
    \item Answer source : Following are different types of sources possible:
    \begin{enumerate}
        \item Image-Span: answer is found verbatim in the image as a contiguous sequence of tokens in the reading order
        \item Question-Span: similar to Image-span but found from question.
        \item Multi-Span: formed by multiple text pieces where each `piece' is found verbatim in the image as contiguous sequence of text tokens (i.e, each piece is an Image-span).
        \item Non-extractive: answer is a numerical answer and is not found verbatim on the text on the image or the question.
    \end{enumerate}
    
    \item Evidence : Following are different types of evidences possible:
     \begin{enumerate}
        \item Text: answer is derived purely by reading text found in the image.
        \item Table/List: finding answer requires one to understand a tabular or list type structure
        \item Visual/Layout: requires one to look for visual aspects(colors, names of objects etc.) or layout of the image to arrive at the answer.
        \item Figure: requires understanding a figure, a plot, a visualization or a schematic.
        
        \item Map: answer is based on a geographical map
        
    \end{enumerate}
    
    \item Operation : if answering the question requires one of the following discrete operations:
    \begin{enumerate}
        \item Counting: requires to count something
        \item Arithmetic: requires to do basic arithmetic operations (sum\/subtract\/multiply\/divide) to find the answer.
        \item Sorting: requires to find sort numbers or need to compare numbers
   \end{enumerate}
\end{enumerate}

\end{enumerate}
\end{tabularx}
\caption{\textbf{Annotation instructions.}  In addition to the written instructions, workers were trained on the process through webinars. During the webinars, each instruction and annotation policy was explained with the help of multiple examples. This helped the workers get familiarize with the kind of questions and answers that are allowed.}
\label{table:table_annotation_instructions}

\end{table*}

\subsection{Data release}

The dataset will be made publicly available under terms set by our data protection team. The question-answer annotations will be made available under CC-BY license. For research and educational uses, the images in the dataset will be available for download as single zip file. For other uses, a list of original  URLs for all the images in the dataset will be provided.

\section{Additional Statistics and Analysis of\\\datasetname dataset}
\label{appendix:analysis}
In this section we provide additional statistics and analysis of the \datasetname dataset. This section extends Section 3 in the main paper.

\begin{figure*}[b]
    \centering
    \includegraphics[width=0.8\linewidth]{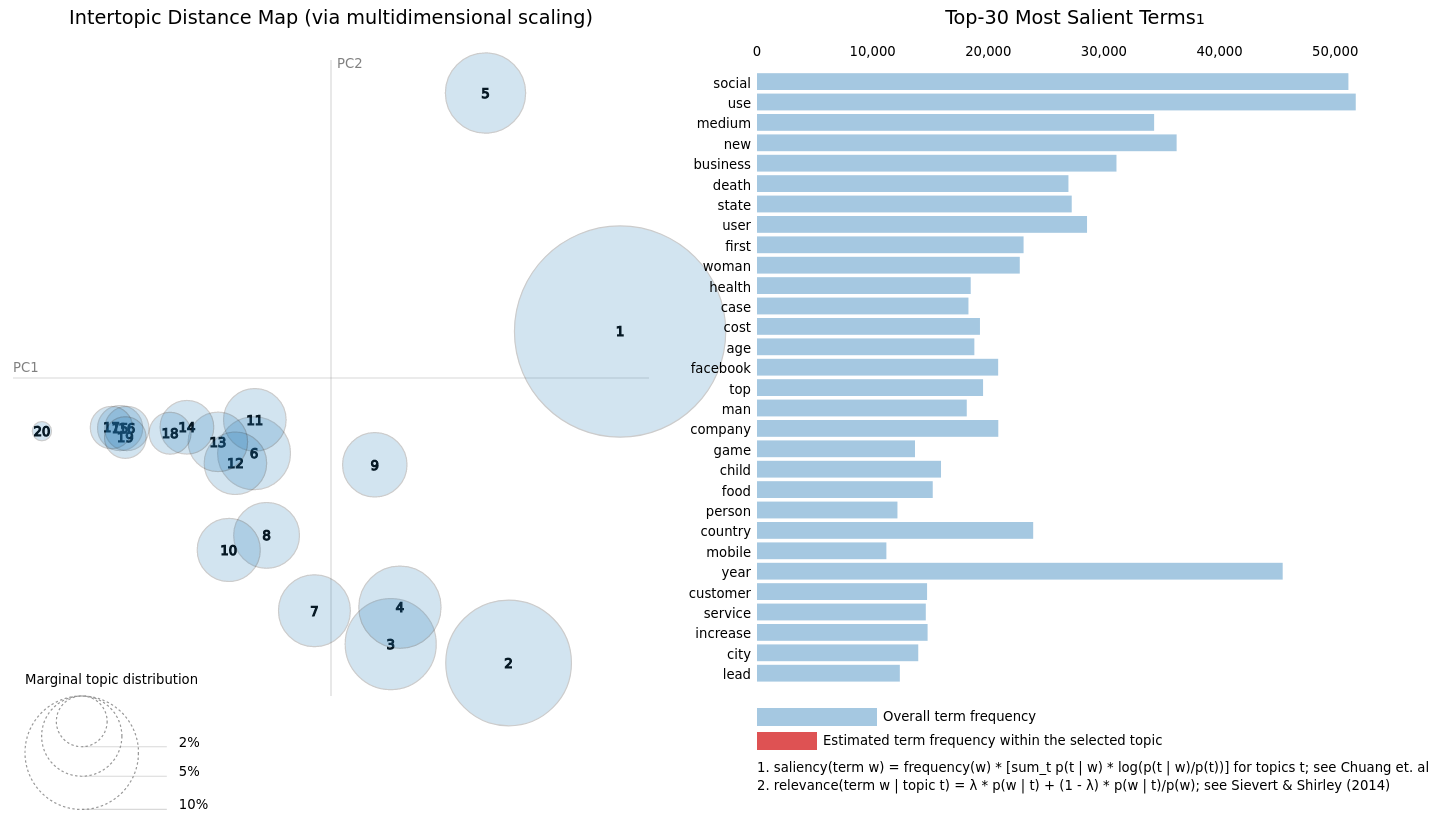}
    \caption{\textbf{Visualization of the top 20 topics in \datasetname dataset.} We used LDA to find the topics. On the left is an inter topic distance map where each circle represent a topic. The area of the circles is proportional to the overall relevance of the topic. Distance between two topics are computed using Jensen–Shannon divergence. The distances are then mapped to two dimensional space using Multidimensional Scaling~\cite{mds}. On the right we show top 30 most salient terms(most prevalent terms in the entire corpus) among the text present on the images. This diagram was created using pyLDAvis tool.}
    \label{fig:topic_model}
\end{figure*}

To analyze the topics covered by images in \datasetname, we used the Latent Dirichlet Allocation (LDA)~\cite{lda} model. We used the LDA implementation in Gensim library~\cite{gensim}. Since our dataset comprises images, the text recognized from the images using an OCR are used for the topic modelling.
\autoref{table:topic_model_table} shows that images in \datasetname dataset cover a wide range of topics such as energy, war, health and social media. In~\autoref{fig:topic_model} we show a visualization of the top 20 topics in the dataset, visualized using the pyLDAvis tool~\cite{pyldavis}.

\begin{figure*}[t]
    \centering
    \includegraphics[width=0.8\linewidth]{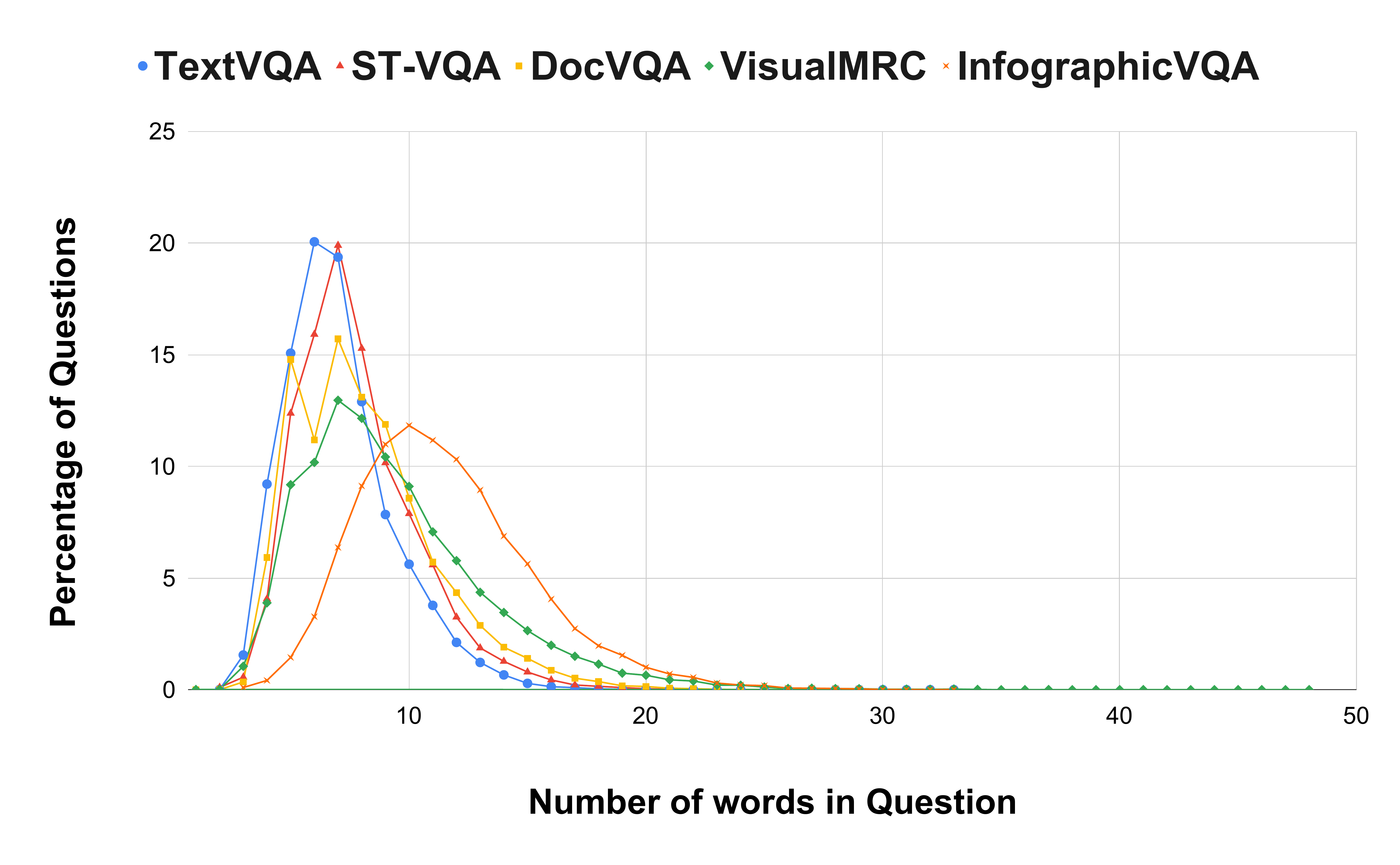}
    \caption{\textbf{Percentage of questions with a particular length.} Compared to other similar datasets, questions in \datasetname are longer on average. Average question length  is 11.54 ( Table 2 in the main paper), which is highest among  similar datasets including VisualMRC, which is an abstractive QA dataset.  }
    \label{fig:questionl}
\end{figure*}

\begin{figure*}[b]
    \centering
    \includegraphics[width=0.8\linewidth]{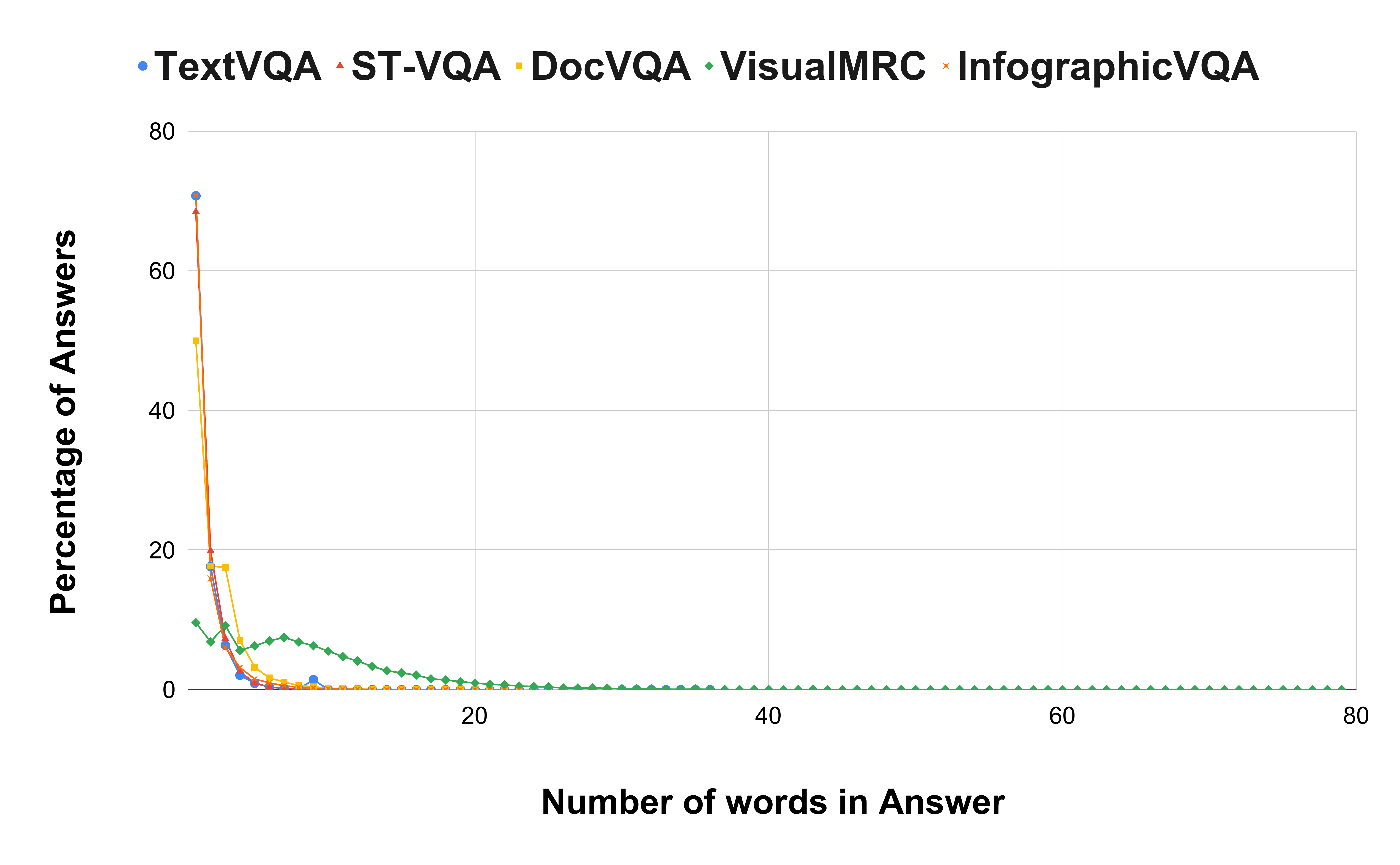}
    \caption{\textbf{Percentage of answers with a particular length.} Answers in \datasetname are shorter compared to most of the similar datasets. More than 70\% of the answers have only single word in it and more than 85\% have at most 2 words in it. This is expected since the questions are asked on data presented on infogrpahics, which is mostly numerical data.  }
    \label{fig:answerl}
\end{figure*}

\begin{figure*}[h]
    \centering
    \includegraphics[width=0.8\linewidth]{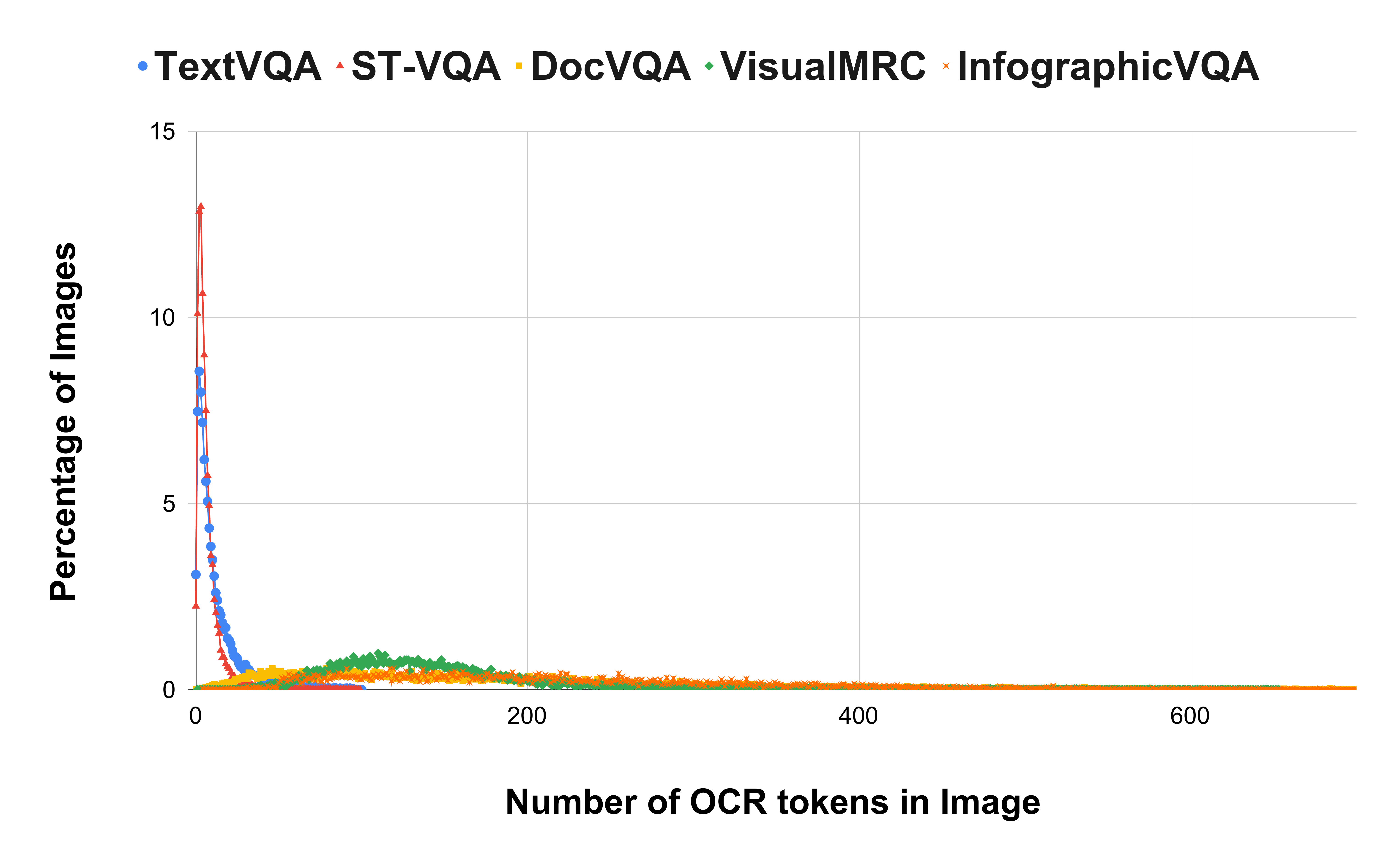}
    \caption{\textbf{Percentage of images with a particular number of text tokens on it.} Average number of text tokens per image is highest in \datasetname (Table 2 in the main paper). It can be seen from the plot that \datasetname has a flatter curve with a longer tail.}
    \label{fig:tokenl}
\end{figure*}

In~\autoref{fig:questionl},  we plot the distribution of question lengths for questions in \datasetname dataset and other similar datasets --- \textvqa~\cite{textvqa}, \stvqa~\cite{stvqa_iccv}, \docvqa~\cite{docvqa} and \visualmrc~\cite{visualmrc}. Similar plots for answer lengths and number of OCR tokens are shown in~\autoref{fig:answerl} and~\autoref{fig:tokenl} respectively.

\begin{figure*}[h]
    \centering
    \includegraphics[width=0.8\linewidth]{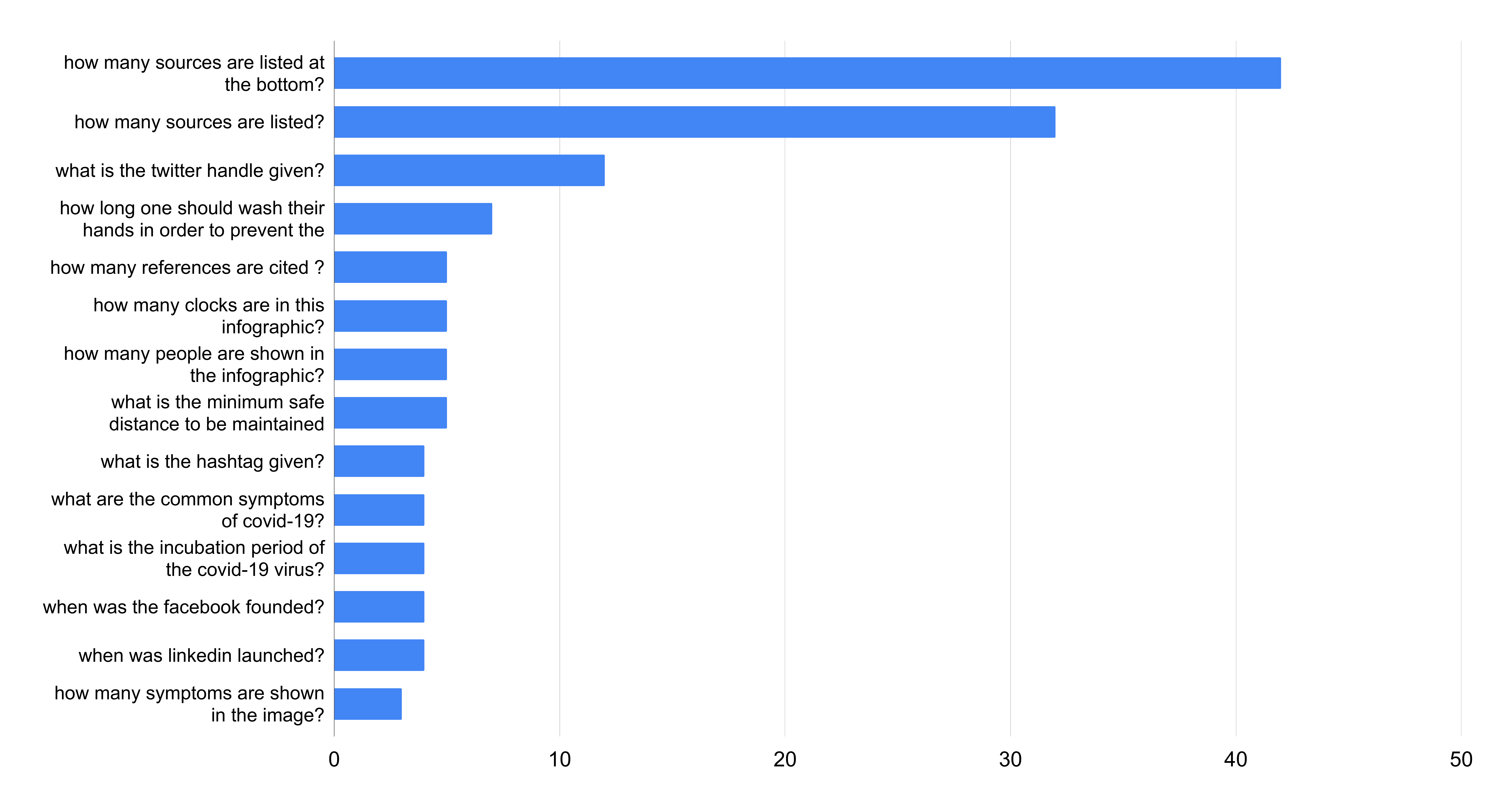}
    \caption{\textbf{Top 15 questions.} A majority of commonly occurring questions concern with counting.  }
    \label{fig:top15_questions}
\end{figure*}

\begin{figure*}[h]
    \centering
    \includegraphics[width=0.8\linewidth]{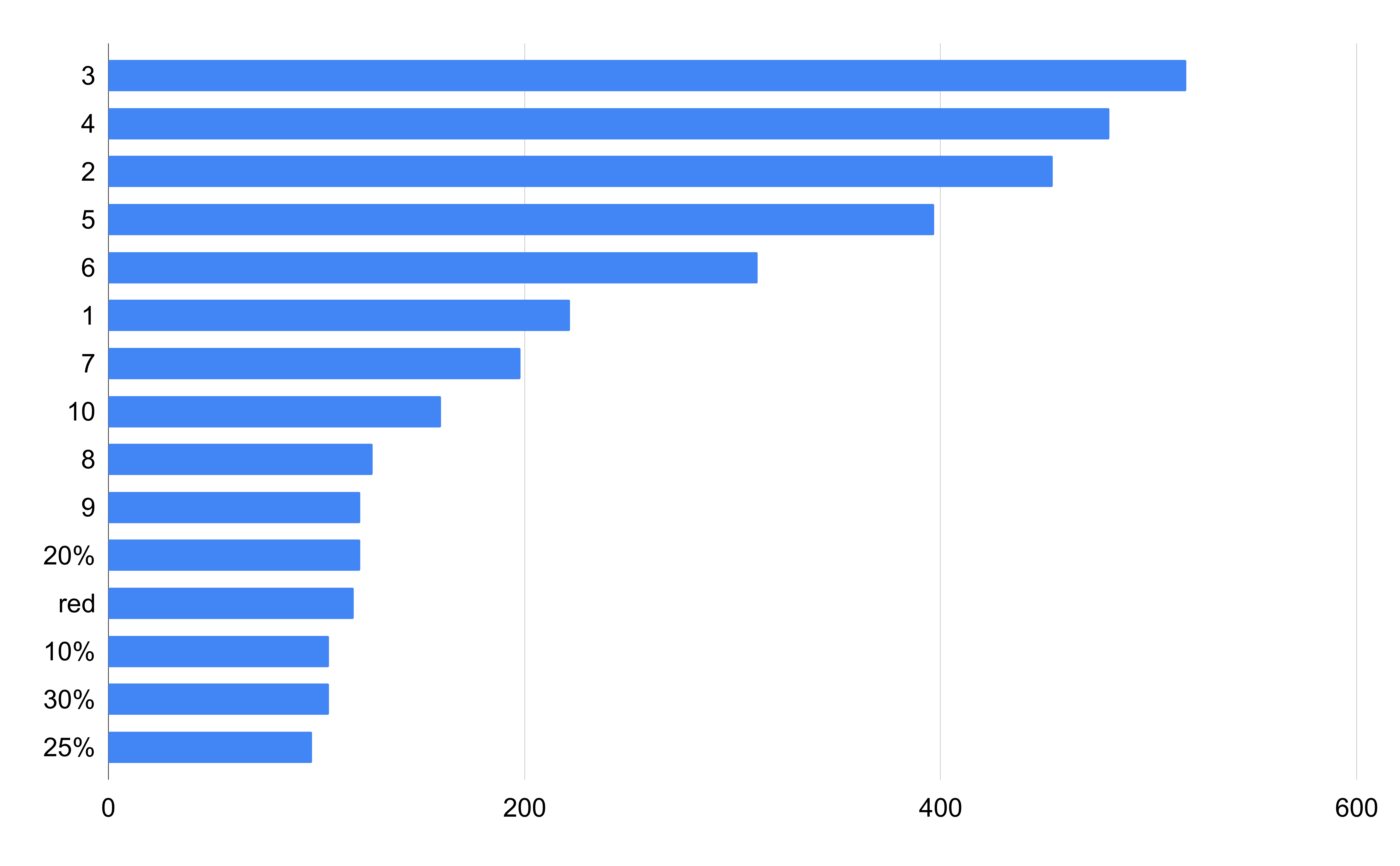}
    \caption{\textbf{Top 15 answers.} Almost all of the top answers are numeric answers.  }
    \label{fig:top15_answers}
\end{figure*}

Top-15 questions in \datasetname dataset based on occurrence frequency are shown in~\autoref{fig:top15_questions}. Top-15 answers and top-15 non-numeric answers are shown in~\autoref{fig:top15_answers} and~\autoref{fig:top15_nonnumeric} respectively.

For \textvqa dataset, for all statistics we use only the publicly available data splits. For statistics and analysis involving OCR tokens, for \datasetname we use OCR tokens spotted by Amazon Textract OCR (we will be making these OCR results publicly available along with our dataset). For \textvqa and \stvqa we use OCR tokens provided as part of data made available in MMF framework~\cite{mmf}. For \docvqa and  \visualmrc we use OCR recognition results  made available as part of the official data releases.

\begin{table*}[b]

\begin{tabularx}{\linewidth}{ll}
\toprule
No. & Topic \\
\midrule
1 & cost lead increase system non risk energy reduce cause clean \\
2 & war violence symptom domestic potential die injury mil acquire birth \\
3 & health person white black police department doctor respiratory smith officer \\
4 & child food water parent potential eat drink essential green sugar already \\
5 & death woman age man old adult love likely statistic rate \\
6 & country high account say month report change global survey event \\
7 & social medium job value program find direct authority salary candidate \\
8 & first purchase call sport still house kid name bring early \\
9 & case university point physical idea language mass brain thought presentation \\
10 & fire act min sunday encounter concentration daily active th monthly \\
11 & paper common check photo add type virus print christmas present \\
12 & game mobile internet app olympic london medal online device mm\_mm \\
13 & public right patient human goal influence earth plant face individual \\
14 & help free american likely provide need support contact tip hand \\
15 & company school design content employee college technology create offer audience \\
16 & new state top city rank york art west east california \\
17 & business customer service population sale product small software increase investment \\
18 & force industry car line waste register decrease driver victim throw \\
19 & year world people day make time com average number source \\
20 & user use facebook share site video post google search worldwide \\
\bottomrule
\end{tabularx}

\caption{\textbf{Top 20 topics in \datasetname found using LDA.} We used text tokens spotted on the images for topic modelling.}
\label{table:topic_model_table}

\end{table*}

\begin{figure*}[h]
    \centering
    \includegraphics[width=0.8\linewidth]{images/top15_nonumeric_answers.pdf}
    \caption{\textbf{Top 15 non-numeric answers.} Non numeric answers are dominated by names of countries, names of colors and names of social media platforms. }
    \label{fig:top15_nonnumeric}
\end{figure*}

\section{More details on experiments}
\label{appendix:experiments}

\subsection{Evaluation}
\label{appendix:evaluation}
Since more than 70\% of the answers in the dataset are taken verbatim from the text present in the image, we decided to keep the evaluation protocol same as the one used for \stvqa~\cite{stvqa_iccv} and \docvqa~\cite{docvqa} benchmarks. These benchmarks where answers are compulsorily extracted from the text on the images, authors propose to use Average Normalized Levenshtein Similarity (ANLS) as the primary evaluation metric. The metric was originally introduced for evaluating VQA on \stvqa. As the authors of \stvqa state, ANLS ``responds softly to answer mismatches due to OCR imperfections''.

The below definition for ANLS is taken  from \stvqa paper.

In~\autoref{eq:lev} $N$ is the total number of questions and  $M$ the number of GT answers per question. $a_{ij}$ are the  the ground truth answers where $i = \{0, ..., N\}$, and $j = \{0, ..., M\}$, and $o_{q_i}$ the predicted answer for the $i^{th}$ question $q_i$. Then, the final score is defined as: %

\begin{equation}
\label{eq:lev}
\textrm{ANLS} = \frac{1}{N} \sum_{i=0}^{N} \left(\max_{j}  s(a_{ij}, o_{q_i}) \right) \\
\end{equation}

\[
  s(a_{ij}, o_{q_i}) =
  \begin{cases}
            (1 - NL(a_{ij}, o_{q_i})) & \text{if \textrm{ $NL(a_{ij},o_{q_i}) < \tau$}} \\
                                    $0$ & \text{if \textrm{ $NL(a_{ij},o_{q_i})$  $\geqslant$  $\tau$}} 
  \end{cases}
\]

\noindent

where $NL(a_{ij}, o_{q_i})$  is the Normalized Levenshtein distance between the lower-cased strings $a_{ij}$ and $o_{q_i}$ (notice that the normalized Levenshtein distance is a value between $0$ and $1$). 
We then define a threshold $\tau = 0.5$ to filter NL values larger than the threshold.
The intuition behind the threshold is that if an output has a normalized edit distance of more than $0.5$ to an answer, it is highly unlikely that the answer mismatch is  due to OCR error.

In addition to ANLS, we also evaluate the performance in terms of Accuracy which is the percentage of questions for which the predicted answer match exactly  with at least one of the ground truth answers. Note that for both the ANLS and Accuracy computation, the ground truth answers and the predicted answers are  converted to lowercase.

\subsection{ Experimental setup for M4C}
\label{appendix:m4c}
This section provides additional details about the experimental setup we used for the M4C~\cite{m4c} model. We used the official implementation available as part of the MMF multimodal learning framework~\cite{mmf}.

We trained our models on 4, NVIDIA RTX 2080Ti GPUs. The maximum number of decoding steps used for the iterative answer prediction module is $12$. The multimodal transformer block had $4$ transformer layers, with $12$ attention heads. The dropout ratio for the transformer block was $0.1$. We used Adam optimizer. The batch size was $128$ and we trained the models for $24,000$ iterations. We used a base learning rate of $1e-04$, a warm-up learning factor of $0.2$ and $2,000$ warm-up iterations. We used a learning rate decay of $0.1$. Gradients were clipped when $L^2$ norm exceeds $0.2$.

\subsection{ Experimental setup for\\\layoutlm}
\label{appendix:layoutlm}
\noindent\textbf{Preparing QA data for finetuning LayoutLM.} We finetune \layoutlm model for SQuAD~\cite{squad} style extractive QA wherein start and end tokens of a span is predicted. For this, we need to prepare \layoutlm training data in SQuAD-style format where answer is marked as a span of the text present on the infographic image. We serialize the OCR tokens spotted on each image in the natural reading order. Then we check if a ground truth answer can be found as a subsequence of the serialized text. In cases where an answer has multiple subsequence matches with the serialized text, the first match is taken as the answer span. If no match is found, the particular question is not used finetuning the model. This approach of finding answer spans from answers is inspired by the similar approach used by authors of TriviaQA~\cite{triviaqa}. The same has been used by authors of DocVQA as well for finetuning a BERT QA model for span prediction. Unlike substring matching in TriviaQA we look for subsequence matches as proposed by DocVQA authors. Substring matches can result in  many false matches. Since \datasetname has lot of numeric answers false matches are even more likely. For example if the answer is ``3'' (the most common answer in \datasetname dataset), if we go by substring matching, it will match with a 3 in `300' and `3m'.

\section{Additional Qualitative Examples}
\label{appendix:Additional Qualitative Examples}
In~\autoref{fig:qual:Image-Span_d1}--~\autoref{fig:qual:multiple_operations_d7}, we show qualitative examples covering multiple types of answers, evidences and operations which we discuss in Section 3.3 of the main paper. These results supplement the qualitative results we show in Section 5.2 in the main paper.

\newcommand{\qualFont}[2]{\footnotesize{\fontfamily{qhv}\selectfont \textbf{#1: } #2}}

\setlength{\tabcolsep}{16pt}  %
\begin{figure*}[h]
\begin{center}
    \begin{tabular}{p{0.25\linewidth}p{0.25\linewidth}}
    \multicolumn{2}{c}{\includegraphics[width=0.5\linewidth,height=0.7\linewidth]{images/supplementary/45014.jpeg}}
    \\
    
    \multicolumn{2}{c}{\fbox{\includegraphics[width=0.3\linewidth,height=0.15\linewidth]{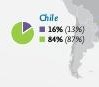}}}
    \\ \\
        \multicolumn{2}{l}{\footnotesize{\fontfamily{qhv}\selectfont \textbf{Q: }what percent of researchers in Chile were men in the duration of 2011-15?}} \\
        
        \qualFont{GT}{\groundtruth{[84\%, 84]}} \par
        \qualFont{\layoutlm}{\wrong{23\%}} \par
        \qualFont{M4C}{\wrong{23\%}} \par
        \qualFont{Human}{\correct{84\%}}
        
        &
        
        \qualFont{Answer-source}{\hspace{0.05cm} \hlc[blue!30]{Image-span}} \par
        \qualFont{Evidence}{\hspace{0.05cm} \hlc[cyan!30]{Map}} \par
        \qualFont{Operation}{\hspace{0.05cm} \textit{none}} \par  \\
        
    \end{tabular}

\end{center}
\caption{\textbf{Using color codes and information on a Map to arrive at answer.} To answer this question, models require to understand that the blue color correspond to women and then pick the number corresponding to the blue color from the data given for Chile. Both the baseline models get this question wrong. Note that here there are two valid answers (GT), one added during the first stage of annotation and the other during the second stage.}
\label{fig:qual:Image-Span_d1}

\end{figure*}

\setlength{\tabcolsep}{6pt}  %

\setlength{\tabcolsep}{16pt}  %
\begin{figure*}[h]
\begin{center}
    \begin{tabular}{p{0.25\linewidth}p{0.25\linewidth}}
    \multicolumn{2}{c}{\includegraphics[width=0.5\linewidth,height=0.8\linewidth]{images/supplementary/10360.jpeg}}
    \\
    
    \multicolumn{2}{c}{\fbox{\includegraphics[width=0.3\linewidth,height=0.2\linewidth]{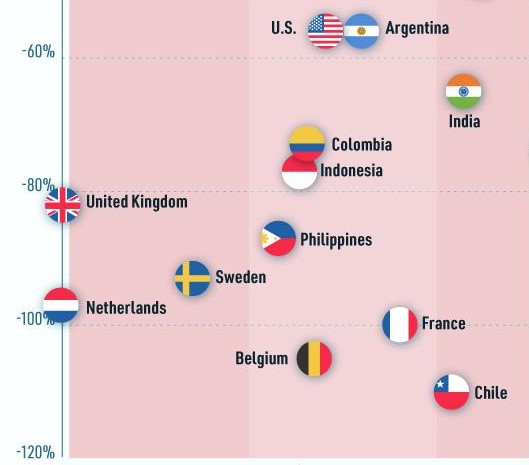}}}
    \\ \\
        \multicolumn{2}{l}{\footnotesize{\fontfamily{qhv}\selectfont \textbf{Q: }Which quadrant does the country India fall into, blue, pink, or gray?}} \\
        
        \qualFont{GT}{\groundtruth{pink}} \par
        \qualFont{\layoutlm}{\wrong{country}} \par
        \qualFont{M4C}{\correct{pink}} \par
        \qualFont{Human}{\correct{pink}}
        
        &
        
        \qualFont{Answer-source}{\hspace{0.05cm} \hlc[blue!30]{Question-span}} \par
        \qualFont{Evidence}{\hspace{0.05cm} \hlc[cyan!30]{Figure} \hlc[cyan!30]{Visual/Layout}}  \par
        \qualFont{Operation}{\hspace{0.05cm} \textit{none}} \par  \\
        
    \end{tabular}

\end{center}
\caption{\textbf{Answer is a color name which is given as a multiple choice option in the question.} To answer this question, a model needs to first locate "India" on the image and then identify the background color there. M4C gets this question correct.}
\label{fig:qual : Multi-Span}

\end{figure*}

\setlength{\tabcolsep}{6pt}  %

\setlength{\tabcolsep}{16pt}  %
\begin{figure*}[h]
\begin{center}
    \begin{tabular}{p{0.25\linewidth}p{0.25\linewidth}}
    \multicolumn{2}{c}{\includegraphics[width=0.5\linewidth]{images/supplementary/39284.jpeg}}
    \\
    
    \multicolumn{2}{c}{\fbox{\includegraphics[width=0.3\linewidth,height=0.2\linewidth]{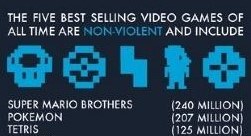}}}
    \\ \\
        \multicolumn{2}{l}{\footnotesize{\fontfamily{qhv}\selectfont \textbf{Q: }Which are the top 2 best selling non violent video games of all time?}} \\
        
        \qualFont{GT}{\groundtruth{super mario brothers, pokemon}} \par
        \qualFont{\layoutlm}{\halfcorrect{super mario brothers}} \par
        \qualFont{M4C}{\wrong{instagram youth}} \par
        \qualFont{Human}{\correct{super mario brothers, pokemon}}
        
        &
        
        \qualFont{Answer-source}{\hspace{0.05cm} \hlc[blue!30]{Multi-span}} \par
        \qualFont{Evidence}{\hspace{0.05cm} \hlc[cyan!30]{Table/List}} \par
        \qualFont{Operation}{\hspace{0.05cm} \hlc[yellow!30]{Sorting}} \par  \\
        
    \end{tabular}

\end{center}
\caption{\textbf{Multi-Span answer.} Multi-span answer type allows us to include questions where answer is formed by multiple single `span's. In this example top 2 items in a category need to be found. This can only be answered if we allow answers containing multiple spans. Since the LayoutLM-based model we train for extractive QA can handle only  single spans, it gets  first part of the answer correct. }
\label{fig:qual_Multi-Span_d3}

\end{figure*}

\setlength{\tabcolsep}{6pt}  %

\setlength{\tabcolsep}{1pt}  %
\begin{figure*}[h]
\begin{center}
    \begin{tabular}{p{0.25\linewidth}p{0.25\linewidth}}
    \multicolumn{2}{c}{\includegraphics[width=0.6\linewidth]{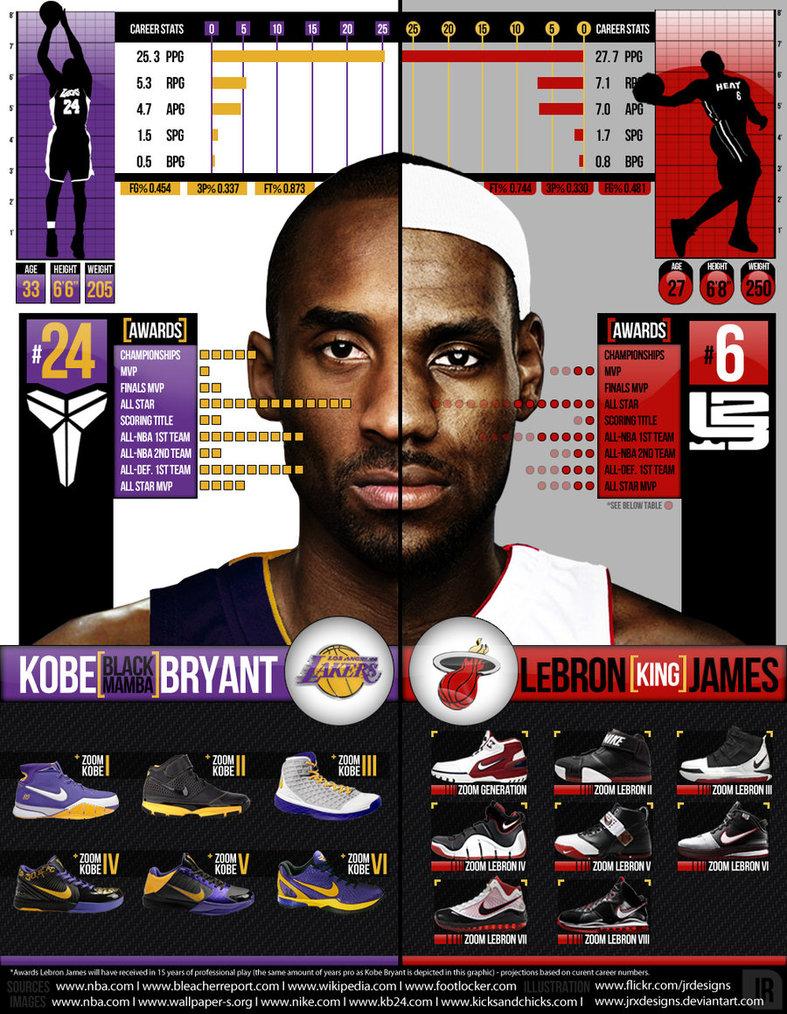}}
    \\
    
    \multicolumn{2}{c}{\fbox{\includegraphics[width=0.3\linewidth,height=0.2\linewidth]{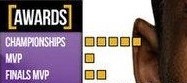}}}
    \\ \\
        \multicolumn{2}{l}{\footnotesize{\fontfamily{qhv}\selectfont \textbf{Q: }How many championships has Kobe Bryant won?}} \\
        
        \qualFont{GT}{\groundtruth{5}} \par
        \qualFont{\layoutlm}{\correct{5}} \par
        \qualFont{M4C}{\correct{5}} \par
        \qualFont{Human}{\correct{5}}
        
        &
        
        \qualFont{Answer-source}{\hspace{0.05cm} \hlc[blue!30]{Non-extractive}} \par
        \qualFont{Evidence}{\hspace{0.05cm} \hlc[cyan!30]{Figure}} \par
        \qualFont{Operation}{\hspace{0.05cm} \hlc[yellow!30]{Counting}} \par  \\
        
    \end{tabular}

\end{center}
\caption{\textbf{Counting symbols/markers to find an answer.} Both the models get the answer correct for  this question that  require one to count the yellow squares next to "CHAMPIONSHIPS". }
\label{fig:qual_Counting_d4}

\end{figure*}

\setlength{\tabcolsep}{6pt}  %

\setlength{\tabcolsep}{16pt}  %
\begin{figure*}[h]
\begin{center}
    \begin{tabular}{p{0.25\linewidth}p{0.25\linewidth}}
    \multicolumn{2}{c}{\includegraphics[width=0.6\linewidth]{images/supplementary/10068.jpeg}}
    \\
    
    \multicolumn{2}{c}{\fbox{\includegraphics[width=0.4\linewidth,height=0.1\linewidth]{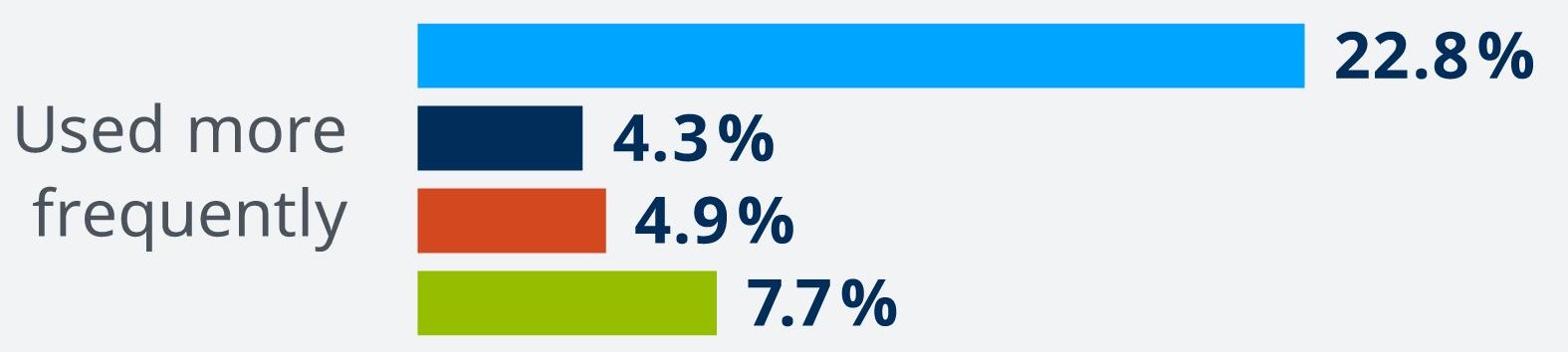}}}
    \\ \\
        \multicolumn{2}{l}{\footnotesize{\fontfamily{qhv}\selectfont \textbf{Q: }Which drug was used more frequently during lockdown, MDMA, Cocaine, Cannabis, or Amphetamines?}} \\
        
        \qualFont{GT}{\groundtruth{cannabis}} \par
        \qualFont{\layoutlm}{\correct{cannabis}} \par
        \qualFont{M4C}{\wrong{cocaine}} \par
        \qualFont{Human}{\correct{cannabis}}
        
        &
        
        \qualFont{Answer-source}{\hspace{0.05cm} \hlc[blue!30]{Question-span} \hlc[blue!30]{Image-span}} \par
        \qualFont{Evidence}{\hspace{0.05cm} \hlc[cyan!30]{Figure}} \par
        \qualFont{Operation}{\hspace{0.05cm} \hlc[yellow!30]{Sorting}} \par  \\
        
    \end{tabular}

\end{center}
\caption{\textbf{Sorting values shown in a bar chart.} In this question, answer is a span of question (Question-span) and a span of the text on the image (Image-span) as well. The largest among the given items is explicit in the bar chart representation. Alternatively the same can be found by finding the largest by comparing the numbers. Hence `Sorting' is added as the Operation.}
\label{fig:qual : Sorting}

\end{figure*}

\setlength{\tabcolsep}{6pt}  %

\setlength{\tabcolsep}{16pt}  %
\begin{figure*}[h]
\begin{center}
    \begin{tabular}{p{0.25\linewidth}p{0.25\linewidth}}
    \multicolumn{2}{c}{\includegraphics[width=0.4\linewidth,height=0.85\linewidth]{images/supplementary/40401.jpeg}}
    \\
    
    \multicolumn{2}{c}{\fbox{\includegraphics[width=0.4\linewidth,height=0.2\linewidth]{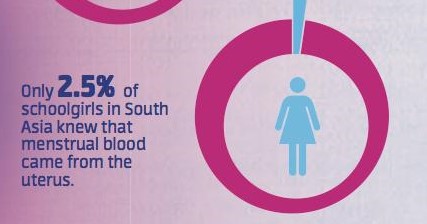}}}
    \\ \\
        \multicolumn{2}{l}{\footnotesize{\fontfamily{qhv}\selectfont \textbf{Q: }What \% of schoolgirls in South Asia do not know that menstrual blood comes from the uterus?}} \\
        
        \qualFont{GT}{\groundtruth{[97.5, 97.5\%]}} \par
        \qualFont{\layoutlm}{\wrong{2.5\%}} \par
        \qualFont{M4C}{\wrong{25}} \par
        \qualFont{Human}{\correct{97.5\%}}
        
        &
        
        \qualFont{Answer-source}{\hspace{0.05cm} \hlc[blue!30]{Non-extractive}} \par
        \qualFont{Evidence}{\hspace{0.05cm} \hlc[cyan!30]{Text}} \par
        \qualFont{Operation}{\hspace{0.05cm} \hlc[yellow!30]{Arithmetic}} \par  \\
    \end{tabular}

\end{center}
\caption{\textbf{Question requiring arithmetic operation.} To answer this question, the given percentage value needs to be subtracted from 100. Both the models fail to get the answer correct.}
\label{fig:qual : Arithmetic}

\end{figure*}

\setlength{\tabcolsep}{6pt}  %

\setlength{\tabcolsep}{16pt}  %
\begin{figure*}[h]
\begin{center}
    \begin{tabular}{p{0.25\linewidth}p{0.25\linewidth}}
    \multicolumn{2}{c}{\includegraphics[width=0.5\linewidth,height=17cm]{images/supplementary/42017.jpeg}}
    \\ \\
    
        \multicolumn{2}{l}{\footnotesize{\fontfamily{qhv}\selectfont \textbf{Q: }Playing against which country did he reach the most number of his milestone runs?}} \\
        
        \qualFont{GT}{\groundtruth{sri lanka}} \par
        \qualFont{\layoutlm}{\wrong{bangladesh}} \par
        \qualFont{M4C}{\wrong{pakistan}} \par
        \qualFont{Human}{\correct{sri lanka}}
        
        &
        
        \qualFont{Answer-source}{\hspace{0.05cm} \hlc[blue!30]{Image-span}} \par
        \qualFont{Evidence}{\hspace{0.05cm} \hlc[cyan!30]{Text}  \hlc[cyan!30]{Figure}} \par
        \qualFont{Operation}{\hspace{0.05cm} \hlc[yellow!30]{Counting} \hlc[yellow!30]{Sorting}} \par  \\
        
    \end{tabular}

\end{center}
\caption{\textbf{Performing  multiple discrete operations.} Here the context required to find the answer spans the entire image. Hence we do not show a crop of the image in the inset. This question requires a model to do Counting --- count number of milestone runs scored against each country and then perform Sorting --- find the country against which the player scored most milestone runs.}
\label{fig:qual:multiple_operations_d7}

\end{figure*}

\setlength{\tabcolsep}{6pt}  %

{\small
\bibliographystyle{ieee_fullname}
\bibliography{references}
}

\end{document}